\documentclass[10pt,twocolumn,letterpaper]{article}

\PassOptionsToPackage{table}{xcolor}
\usepackage{paper}

\usepackage{cuted}
\usepackage{currfile}
\usepackage{caption}

\newcommand{\rom}[1]{\lowercase\expandafter{\romannumeral #1\relax}}

\definecolor{tabfirst}{rgb}{1, 0.7, 0.7}
\definecolor{tabsecond}{rgb}{1, 0.85, 0.7}
\definecolor{tabthird}{rgb}{1, 1, 0.7}
\definecolor{tabgrad}{rgb}{0.25, 0.75, 0.35}

\usepackage{multirow}
\usepackage{pgfplots}
\pgfplotsset{compat=1.18}
\usepgfplotslibrary{groupplots}

\usepackage{algorithm}
\usepackage{algpseudocode}

\definecolor{linkblue}{rgb}{0.21,0.49,0.74}
\usepackage[pagebackref,breaklinks,colorlinks,allcolors=linkblue]{hyperref}

\title{SHrank: A Memory-Efficient Optimizer for Spherical Harmonics in 3D Gaussian Splatting}

\author{Rahul Goel \quad Ahmed H. Mahmoud \quad Jonathan Ragan-Kelley \quad Fr\'edo Durand\\
MIT CSAIL}

\begin{document}
\maketitle

{\renewcommand{\thefootnote}{}\footnotetext{\hspace{-1.5em}\raggedright%
Project page: \url{https://rahul-goel.github.io/shrank} }}

\begin{abstract}

  We propose a memory-efficient optimizer for training 3D Gaussian Splatting
  that preserves the underlying scene representation and reconstruction quality.
  We target the memory overhead of Adam when optimizing Spherical Harmonics,
  which account for a substantial fraction of training memory. By dropping the
  first moment, pooling the second moment across a Gaussian, and exploiting the
  low-rank structure of gradients, we achieve a $3.3\times$ reduction in static
  memory, and a $1.7\times$ reduction in peak memory on standard benchmarks that
  grows to $2.5\times$ at scale. Our approach also lowers the memory bandwidth
  for the optimizer, resulting in $2.5\times$ faster training at scale, all
  while matching the quality of Adam. This allows fitting up to $2.5\times$ more
  primitives on the same GPU, unlocking large scale scene training with fewer
  resources. Our optimizer is a drop-in replacement and cleanly integrates into
  various training systems. In CPU-offloaded training, it can reduce the PCIe
  traffic by $3.7\times$ and the CPU RAM requirement by $2.6\times$. In
  multi-GPU training, it reduces the memory per GPU by up to $1.7\times$.

\end{abstract}
\colorlet{cropred}{linkblue!35}
\begin{figure}[t]
\centering
\newcommand{\teaserdir}{figures/teaser_img}%
\newcommand{\teasercropx}{721}\newcommand{\teasercropy}{171}\newcommand{\teasercrops}{416}%
\begin{tikzpicture}
  \node[anchor=south west, inner sep=0] (img) {\includegraphics[width=\linewidth]{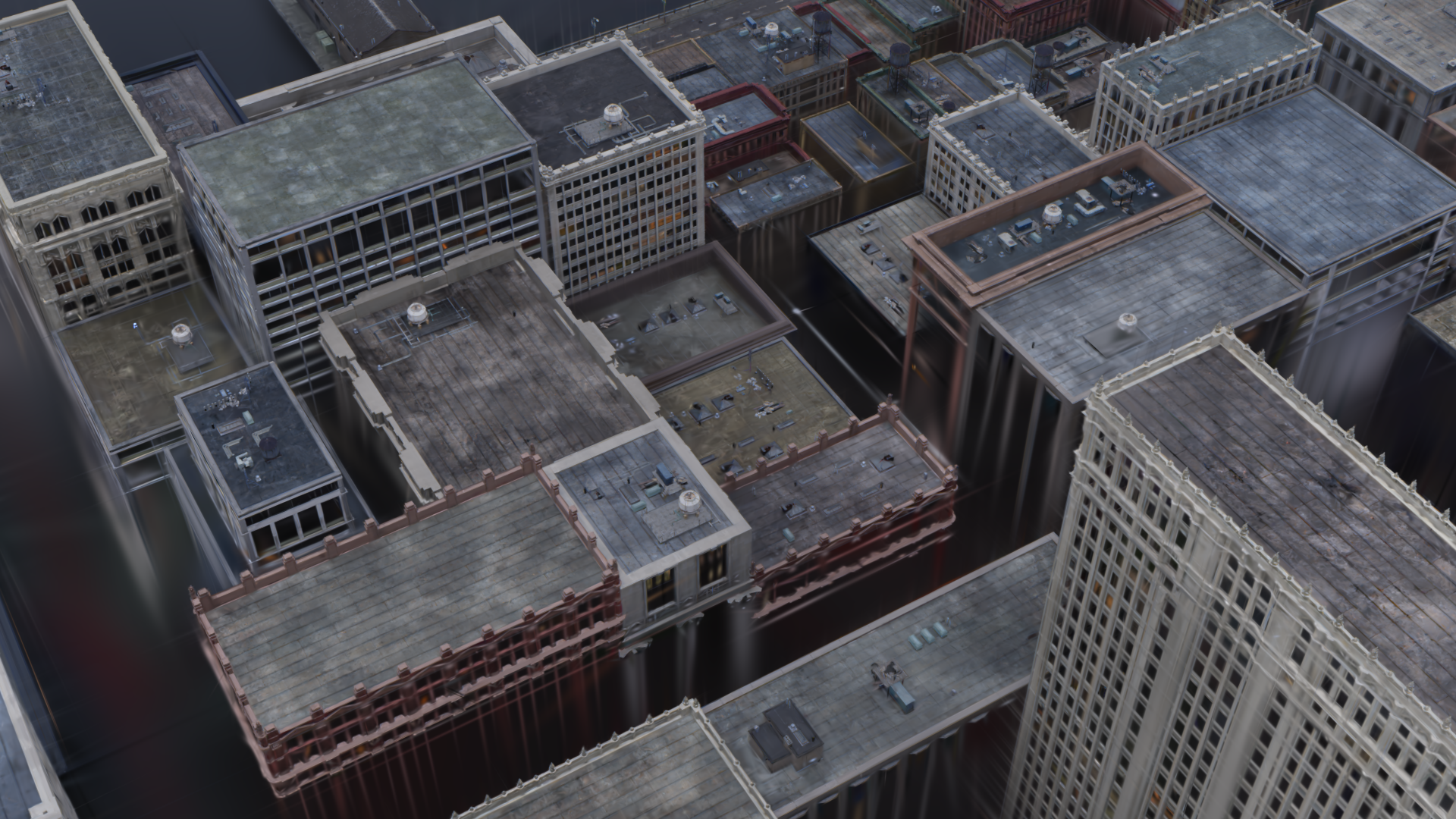}};
  \begin{scope}[x={(img.south east)}, y={(img.north west)}]
    \draw[cropred, line width=0.8pt] (\teasercropx/1920, 1-\teasercropy/1080)
      rectangle ({(\teasercropx+\teasercrops)/1920}, {1-(\teasercropy+\teasercrops)/1080});
  \end{scope}
\end{tikzpicture}\\[1pt]
\setlength{\tabcolsep}{1pt}\setlength{\fboxsep}{0pt}\setlength{\fboxrule}{0.6pt}%
\newlength{\teasercol}\setlength{\teasercol}{\dimexpr(\linewidth-4pt)/3\relax}%
\newcommand{\teasercrop}[1]{\includegraphics[width=\teasercol]{\teaserdir/crops/#1_crop.png}}%
{\scriptsize\renewcommand{\arraystretch}{1.0}%
\begin{tabular}{@{}*{3}{>{\centering\arraybackslash}p{\teasercol}}@{}}
  Ground truth & {[Adam + bwd.\ fusion]} & {[Ours + bwd.\ fusion]} \\
  \teasercrop{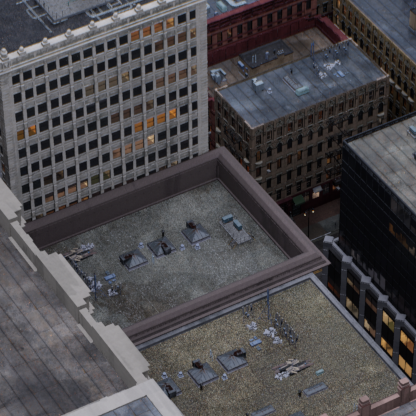} & \teasercrop{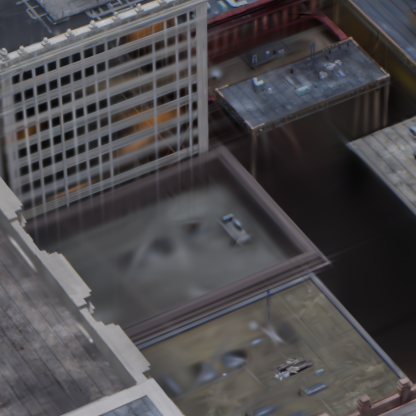} & \teasercrop{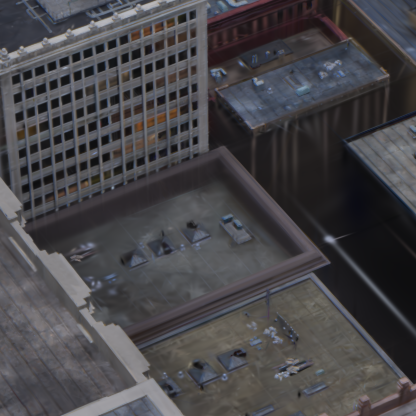} \\
  \#G\,/\,Peak VRAM, Time & 10.0M\,/\,10.2\,GB, 4.7\,h & 25.3M\,/\,10.4\,GB, 4.2\,h \\
  PSNR\,/\,SSIM\,/\,LPIPS & 24.41\,/\,0.734\,/\,0.378 & 25.16\,/\,0.766\,/\,0.322 \\
\end{tabular}}
\caption{When training with equal available memory (12\,GB TITAN Xp), our
method allows fitting $2.5 \times$ more Gaussians in less training time, and the
higher primitive count leads to a higher reconstruction quality.  Both methods
start from the same initialization on MatrixCity (small city, aerial views).
Averaged over all test views, our method reaches $25.15$\,dB PSNR compared to
$23.78$\,dB for Adam.  Metrics below the crops are for the shown view.}
\label{fig:teaser}
\end{figure}

\section{Introduction}
\label{sec:introduction}

The seminal work of 3D Gaussian Splatting (3DGS)~\cite{kerbl3Dgaussians} has
changed the field of 3D reconstruction and novel view synthesis. It represents
the scene as explicit fuzzy 3D Gaussians, and performs differentiable
rasterization on them efficiently for rendering and training. 3DGS can
reconstruct scenes ranging from small ones~\cite{kerbl3Dgaussians} to large
scale cities~\cite{grendelgs,zhao2025clm,zhong2026tidegs}. For training, 3DGS
uses Adam~\cite{kingma2015adam}---which maintains two additional states for each
parameter making it the primary occupant of GPU memory. For large scenes, this
inhibits large primitive counts and leads to poorer reconstructions as shown in
Figure~\ref{fig:teaser}.

In both deep learning and 3DGS, Adam's high memory hinders scalability and
finding a new memory-efficient optimizer with similar convergence properties is
challenging due to the loss landscape. New optimizers are gaining traction in
deep learning but they target neural networks with 2D weight matrices rather
than graphics primitives.

In 3DGS, each Gaussian stores $48$ Spherical Harmonics (SH) coefficients out of
a total of $59$ parameters, and Adam keeps two extra states for every parameter.
Hence, the SH optimizer states alone take 40\% of the per-Gaussian training
memory, making them our primary target.

\begin{figure}
  \centering
  \includegraphics[width=0.40\textwidth]{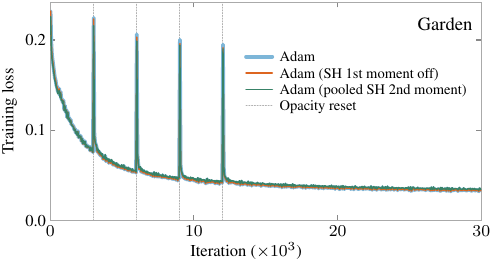}
   \caption{
   With $\beta_1=0$ for Spherical Harmonics (SH), and with mean-pooling the
   second moment for SH, we see negligible impact on the training loss
   trajectory compared to Adam. Similar behavior is seen on other scenes as
   shown in the supplementary.
   }

   \label{fig:garden_train_loss}
\end{figure}

Our starting point is the observation that momentum does not always improve
stochastic optimization~\cite{Kidambi:2018:OTI,Yan:2018:AUA}. In particular,
Wang et al.~\cite{Wang:2024:TMV} show that, for sufficiently small learning
rates, SGD with and without momentum can exhibit similar training behavior.
Their analysis does not directly apply to Adam, but since SH uses very small
learning rate in 3DGS training, it raises the question: \emph{do SH coefficients
in 3DGS benefit from Adam's first-moment estimate?} Similarly,
Adam-mini~\cite{zhang2025adammini} shows that Adam can retain its effectiveness
by sharing a single second moment within blocks of parameters.  We find that
both setting $\beta_1=0$ and sharing a single second moment across all SH
coefficients of a Gaussian have almost no effect on the training loss
(Figure~\ref{fig:garden_train_loss}), while together reducing the SH optimizer
state from $96$ values to just $1$ per Gaussian.

The more interesting structure lies in the SH gradient itself. Due to the way
3DGS evaluates SH, its gradient is an outer product of the RGB color gradient
and the SH basis, i.e., a rank-$1$ matrix. This enables us to reconstruct the
gradient tensor on-the-fly, removing another major memory-consuming buffer.
Furthermore, it naturally pairs with the pooled second moment through Uns\"old's
theorem~\cite{unsoldtheorem}, as the per-step pooled squared gradient comes out
to be the product of the squared norm of the RGB gradient and a constant
independent of the view direction (\S\ref{sec:method_fused_kernel}).

Building on these observations, we propose a memory-efficient optimizer derived
from Adam whose SH update runs in a single fused kernel inside the backward
pass. We call it \emph{SHrank} after the rank-$1$ structure of the SH gradient
and the shrinking effect it has on the optimizer states. SHrank achieves a
$5.1\times$ reduction in optimizer state per Gaussian, a $3.3\times$ smaller
static memory allocation, and $1.7\times$ lower peak training memory on standard
benchmarks, reaching $2.5\times$ at scale and allowing up to $2.5\times$ more
primitives on the same GPU. The reduced memory traffic leads to $1.5\times$
faster end-to-end training that grows to $2.5\times$ at scale.

We evaluate our method across datasets, and experiment with densification
schemes and training implementations. We retain high quality reconstruction and
improve performance with scale. We also compare against many existing first
order optimizers and show that our optimizer matches their quality at a lower
memory and training time.

\section{Related Work}
\label{sec:related-work}
We focus our discussion on prior work relevant to reducing the memory
requirements of 3DGS during training, i.e., methods for scaling to larger
scenes, compressing the representation, and reducing optimizer memory. We also
briefly review work on accelerating the training pipeline.

\subsection{Scaling to Larger Scenes}
\label{sec:rw_large_scenes}
Scaling 3DGS to larger scenes is limited by the memory required for the
primitives and their optimizer states. We classify works on scaling into two
categories. \textit{Representation-Preserving} methods keep the 3DGS
representation unchanged, while \textit{Representation-Altering} methods either
change the representation or train only a part of the scene.

\textit{Representation-Preserving:} These methods keep the same image formation
model as original 3DGS, but change the system around it to accommodate more
primitives. Grendel~\cite{grendelgs} distributes Gaussians and their
optimizer states across multiple GPUs and uses sparse all-to-all communication
to send each GPU the Gaussians needed to render its share of pixels.
CLM~\cite{zhao2025clm} trains on a single GPU by keeping only the geometry of
all the Gaussians on the GPU for quick frustum culling, while offloading the SH
and their optimization to the CPU. CLM trains up to 102 M Gaussians on a 24 GB
GPU. TideGS~\cite{zhong2026tidegs} extends the memory hierarchy to the SSD and
keeps only a working set of Gaussians on the GPU. To avoid moving optimizer
states across the memory hierarchy, it discards the Adam moments of the
Gaussians evicted from the GPU and re-initializes them to zero, pushing training
to a billion Gaussians on a single $24$ GB GPU.  In all three systems, the
gradients and Adam moments are a large part of what has to be sharded,
transferred, or discarded. Our method reduces this state directly, and we
integrate it into Grendel and CLM (see \S\ref{sec:results_other_systems}).

\textit{Representation-Altering:} Divide-and-conquer
methods~\cite{hierarchicalgaussians24} split the scene into chunks and train
them independently and merge them later. Level-of-detail
methods~\cite{renoctreegs,windisch2026ALoDOfGaussians} organize the primitives
into a hierarchy and render a different subset depending on the distance to the
camera. This direction is orthogonal to ours, and our optimizer can be
incorporated into these representations as well.

\subsection{Faster Training}
\label{sec:rw_training_time}

The training pipeline for 3DGS has undergone several performance optimizations.
StopThePop~\cite{radl2024stopthepop} improves the estimation of overlapped tiles
for each projected 2D Gaussian. Taming3DGS~\cite{10.1145/3680528.3687694}
improves the speed of gradient backpropagation from pixels to 2D splats by
reorganizing the parallelization scheme and introduces an efficient
implementation of SSIM loss calculation. Recently,
Faster-GS~\cite{hahlbohm2026fastergs} has evaluated, improved, and bundled the
best available optimizations into a new implementation. We treat Faster-GS as
the fastest publicly available training pipeline that is faithful to the
original 3DGS~\cite{kerbl3Dgaussians} implementation. Other
methods~\cite{fastgs,sadgs} improve training time by changing the
number of primitives and pixel overdraw statistics, and are orthogonal to our
work. Most of these optimizations target the rasterization and the loss, and
leave the optimizer state untouched. As the rest of the pipeline gets faster,
the optimizer takes up a larger share (Amdahl's law), which is our target in
this paper.

\subsection{Compressing Spherical Harmonics}
\label{sec:rw_sh_compression}

SH are the primary target of 3DGS compression literature as they take up the bulk of representation memory~\cite{3DGSzip2025}.
LightGaussian~\cite{lightgaussian} distills the higher order SH bands into lower
ones for a trained scene and quantizes the remaining attributes.
\citet{PapantonakisReduced3dgs} adaptively choose SH bands count for
each Gaussian and store the attributes using codebooks and half-float precision.
Although some of these reductions are applied during training, they occur after peak memory usage and thus do not reduce training memory requirements.
These works target the storage and rendering of a converged representation. In contrast, our work targets the training and can complement post-training techniques.

\subsection{Optimizers}
\label{sec:rw_optimizers}

\paragraph{General Memory-Efficient Optimizers.}
With the scaling of deep learning models, the community has been seeking
replacements for Adam to mitigate its high memory
demand.  RMSProp~\cite{tieleman2012rmsprop} predates Adam and keeps only the
second moment, while Adafactor~\cite{pmlr-v80-shazeer18a} factorizes the
second moment of weight matrices into row and column vectors.
Adam-mini~\cite{zhang2025adammini} partitions parameters into blocks along parameter boundaries and keeps a single second moment across each block.
8-bit Adam~\cite{dettmers2022optimizers} tracks both moments but block-quantizes
them to a byte. VectorAdam~\cite{Selena:2022:VFR} also pools the second moment
across a vector to preserve rotation equivariance. Our work is inspired by these previous works and also exploits the structure of SH coefficients in 3DGS to achieve a memory-efficient optimizer.

\paragraph{3DGS Optimization.}
Taming3DGS~\cite{10.1145/3680528.3687694} introduces
\textit{Reduced SH Adam}, which accumulates higher order SH gradients over 16
iterations before updating; and \textit{Selective Adam}, which only updates the
Gaussian primitives visible in the camera view frustum. The Fused Adam of
Faster-GS~\cite{hahlbohm2026fastergs} performs the Adam step of visible
Gaussians inside the backward pass (we refer to as \emph{backward fusion}).
These methods reduce global memory I/O leading to faster training
time without changing Adam optimizer states.
Our work
shrinks the optimizer states which reduces memory requirement. As a result, we reduce global memory I/O and training time.  3DGS Second-order
methods~\cite{3dgs2,hoellein_2025_3dgslm,second_order_residual,hsiao2026dgstr}
approximate the loss curvature to improve optimization, but introduce additional memory and time costs, and are often incompatible with densification.

\section{Preliminaries}
\label{sec:pre}

This section introduces the concepts underlying our contributions towards memory-efficient optimizer for 3DGS.

\subsection{Scene Representation}
\label{sec:pre_representation}
3DGS models the scene as a set of unnormalized 3D Gaussian primitives,
each with a position $\mu$, scale $s$, rotation $q$, opacity $o$, and Spherical
Harmonics (SH) coefficients for its view-dependent color. SH are orthonormal
basis functions on the unit sphere built from the associated Legendre
polynomials~\cite{sloan2008stupid}, that are organized into bands $l$ of $2l+1$
functions and indexed by the order $m$. The color of a Gaussian from the viewing
direction $d$ is a weighted sum over the $L+1$ bands:
\begin{equation}
c(d) = \sum_{l=0}^{L} \sum_{m=-l}^{l} k_l^m\, Y_l^m(d), \qquad k_l^m \in \mathbb{R}^3,
\label{eq:sh_double_sum}
\end{equation}
where $k_l^m$ holds the RGB coefficients of the basis function $Y_l^m$, and $d$
is the viewing direction. Band $0$ is independent of direction and gives the
view-independent diffuse color (also called \emph{DC}). The higher bands
model view-dependent specular effects.  3DGS uses $L = 3$, i.e., $(L+1)^2 = 16$
basis functions resulting in $48$ coefficients per Gaussian.

\subsection{Rendering}
\label{sec:pre_rendering}
For a given camera, the 3D Gaussians are transformed to the camera coordinate
system and culled with the view-frustum's coverage. The remaining primitives are
projected to 2D~\cite{ewa_splatting}, depth-sorted, and $\alpha$-blended with
the color coming from the evaluation of the SH~\cite{kerbl3Dgaussians}.

However, the authors of 3DGS make an approximation in
SH evaluation. They assumed that the Gaussian has a
small pixel footprint and that the pixel-wise variation in the specular
appearance of each Gaussian is negligible. Therefore, they evaluate the SH
basis once for a given camera---the vector $d$ joining the $3$D
camera center to the $3$D mean of the Gaussian provides the direction of SH
evaluation, which remains constant across pixels for a given view. Flattening
$(l, m)$ into a single index, Equation~\ref{eq:sh_double_sum}
becomes \begin{equation} c(d) = K\,Y(d), \quad K \in \mathbb{R}^{3\times16},\;
Y(d) \in \mathbb{R}^{16}, \label{eq:sh_color} \end{equation} where $K$ is the
SH coefficients of the three color channels and $Y(d)$ is the SH basis evaluated
at $d$. Avoiding the per-pixel calculation keeps the SH evaluation cheap.  We omit the offset and clamp of the rasterized color, $\max(0, 0.5 +
c)$~\cite{kerbl3Dgaussians}; all color gradients $\frac{\partial L}{\partial c}$
are taken before clamping.

\subsection{Adam Optimizer}
\label{sec:pre_adam}
3DGS uses the Adam optimizer~\cite{kingma2015adam} to train all of its
parameters. The primary idea of Adam is to perform temporal filtering on the
noisy gradients received due to stochastic gradient descent using exponentially
moving averages (EMA).  Adam maintains two running moments of the gradient $g$
of the parameters: (\rom{1}) the first moment $m$ maintains an EMA of the
gradient with a factor of $\beta_1$, (\rom{2}) the second moment $\nu$ maintains
an EMA of the square of the gradient with a factor of $\beta_2$. The modified
parameters update rule becomes (omitting the bias correction and the $\epsilon$
term for brevity; all operations are element-wise):
\begin{equation}
\begin{aligned}
m_{t+1} &= \beta_1 m_t + (1 - \beta_1)\, g_t, \quad
\nu_{t+1} &= \beta_2 \nu_t + (1 - \beta_2)\, g_t^2, \\
\theta_{t+1} &= \theta_t - \alpha\, m_{t+1} / \sqrt{\nu_{t+1}}.
\end{aligned}
\label{eq:adam}
\end{equation}

\subsection{Memory Footprint of Training}
\label{sec:pre_memory}
\begin{figure}[t]
\centering
\providecolor{linkblue}{rgb}{0.21,0.49,0.74}
\begin{tikzpicture}[
  font=\scriptsize,
  box/.style={draw=black!60},
  gone/.style={draw=black!40, densely dashed},
  op/.style={font=\small},
  seg/.style={font=\tiny},
]
\colorlet{thS}{linkblue!35}         \colorlet{thN}{linkblue!12}
\colorlet{gS}{orange!45}           \colorlet{gN}{orange!15}
\colorlet{mS}{green!45!black!30}   \colorlet{mN}{green!45!black!10}
\colorlet{vS}{violet!35}           \colorlet{vN}{violet!12}
\def\h{0.3}
\def\tx#1{3.4+0.0041*(#1)}
\def\seg#1#2#3#4#5#6{%
  \fill[#4] ({\tx{#1}},0) rectangle ({\tx{#1+#2}},-\h);
  \fill[#5] ({\tx{#1+#2}},0) rectangle ({\tx{#1+#2+#3}},-\h);
  \draw[box] ({\tx{#1}},0) rectangle ({\tx{#1+#2+#3}},-\h);
  \node[seg] at ({\tx{#1+(#2+#3)/2}},-\h/2) {#6};
}

\def\lgitem#1#2{\tikz[baseline=-0.5ex]\fill[#1] (0,-0.06) rectangle (0.12,0.06);\hspace{0.04cm}#2}
\node[seg, anchor=south west, inner sep=0pt] at ({\tx{0}},0.05)
  {\makebox[3.87cm][s]{\lgitem{thS}{Primal}\hfill\lgitem{gS}{Gradient}\hfill
   \lgitem{mS}{1st Moment}\hfill\lgitem{vS}{2nd Moment}}};

\fill[black!12] (0.35,0) rectangle (0.6856,-\h);
\fill[black!40] (0.6856,0) rectangle (2.15,-\h);
\draw[box] (0.35,0) rectangle (2.15,-\h);
\draw[box] (0.6856,0) -- (0.6856,-\h);
\node[seg] at (0.5178,-\h/2) {$11$};
\node[seg] at (1.4178,-\h/2) {$48$ SH};
\node[op] at (1.25,-0.45) {$\times$};
\foreach \x/\c in {0.35/thS, 0.8/gS, 1.25/mS, 1.7/vS}{
  \draw[box, fill=\c] (\x,-0.6) rectangle ++(0.45,-\h);
  \node[seg] at (\x+0.225,-0.75) {$4$\,B};}
\draw[black!65, densely dotted, line width=0.6pt, rounded corners=2pt] (0.27,0.08) rectangle (2.23,-0.98);

\begin{scope}[yshift=0cm]
\node[anchor=east] (adam) at (3.36,-\h/2) {Adam};
\draw[-stealth, black!70] (2.29,-\h/2) -- ([xshift=2pt]adam.west);
\foreach \i/\n/\s in {0/thN/thS, 1/gN/gS, 2/mN/mS, 3/vN/vS}{
  \seg{236*\i}{44}{192}{\n}{\s}{$236$}}
\node[anchor=west] (adamtot) at ({\tx{944}},-\h/2) {\textbf{$\boldsymbol{944}$\,B}};
\end{scope}

\ifdefined\memoryonlyadam\else
\begin{scope}[yshift=-0.6cm]
\node[anchor=east] at (3.36,-\h/2) {Ours};
\seg{0}{44}{102}{thN}{thS}{$146$}
\seg{146}{44}{0}{gN}{gS}{}
\seg{190}{44}{0}{mN}{mS}{}
\seg{234}{44}{4}{vN}{vS}{}
\foreach \b/\d/\l in {168/-1/44, 212/0/44, 258/1/48}{
  \draw[black!50] ({\tx{\b}},-\h) -- ({\tx{212}+0.34*\d},-\h-0.16);
  \node[seg, anchor=north, inner sep=1pt] at ({\tx{212}+0.34*\d},-\h-0.16) {$\l$};}
\node[anchor=west] (tot) at ({\tx{282}},-\h/2) {\textbf{$\boldsymbol{282}$\,B}};
\node[font=\footnotesize\bfseries, text=green!40!black, anchor=east]
  at (adamtot.east |- tot)
  {$\mathbf{3.3\times}$ smaller};
\end{scope}
\fi
\end{tikzpicture}
\caption{Static training memory per Gaussian in bytes. Each parameter keeps
$4$ buffers under Adam, of which the $48$ SH coefficients (darker shades)
take $81\%$. Our method shrinks only their buffers; the other $11$ parameters
(lighter shades) stay under Adam.}
\label{fig:memory}
\end{figure}
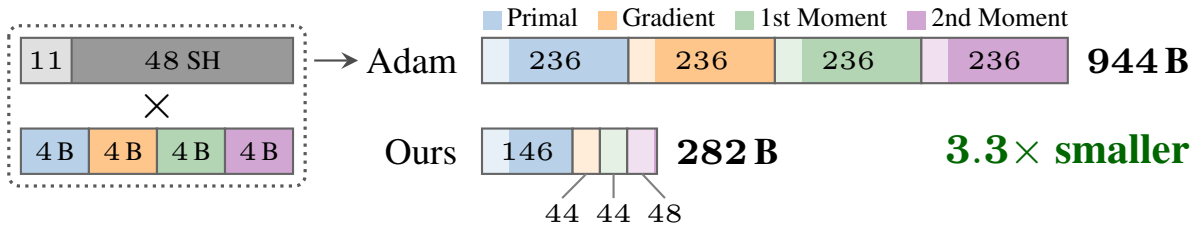

SH carry the bulk of the parameters of a 3D Gaussian---$48$ out
of $59$ (about $81\%$). In 3DGS training, additional memory is allocated for the gradients, and the first and the second moments of
Adam~\cite{kingma2015adam}. Thus, for training $N$
Gaussians, at least $59 \times N \times 4 \times 4$ bytes of static memory is
required, out of which SH take the same proportion in each of
the four categories (primal, gradient, and the two moments), making them the
majority of the memory allocation and bandwidth (Figure~\ref{fig:memory}, top).
In addition, for each training iteration, additional memory is allocated
on-the-fly for the intermediate projected 2D Gaussians and their gradients.
We refer to the former as \emph{static} memory, as it scales with $N$
and persists across iterations, and to the latter as \emph{dynamic} memory, as
it scales with the number of visible Gaussians and the image resolution and is
freed after every iteration.

\section{Method}
\label{sec:method}
\providecommand{\methodsec}[2]{\S\ref{#1}}
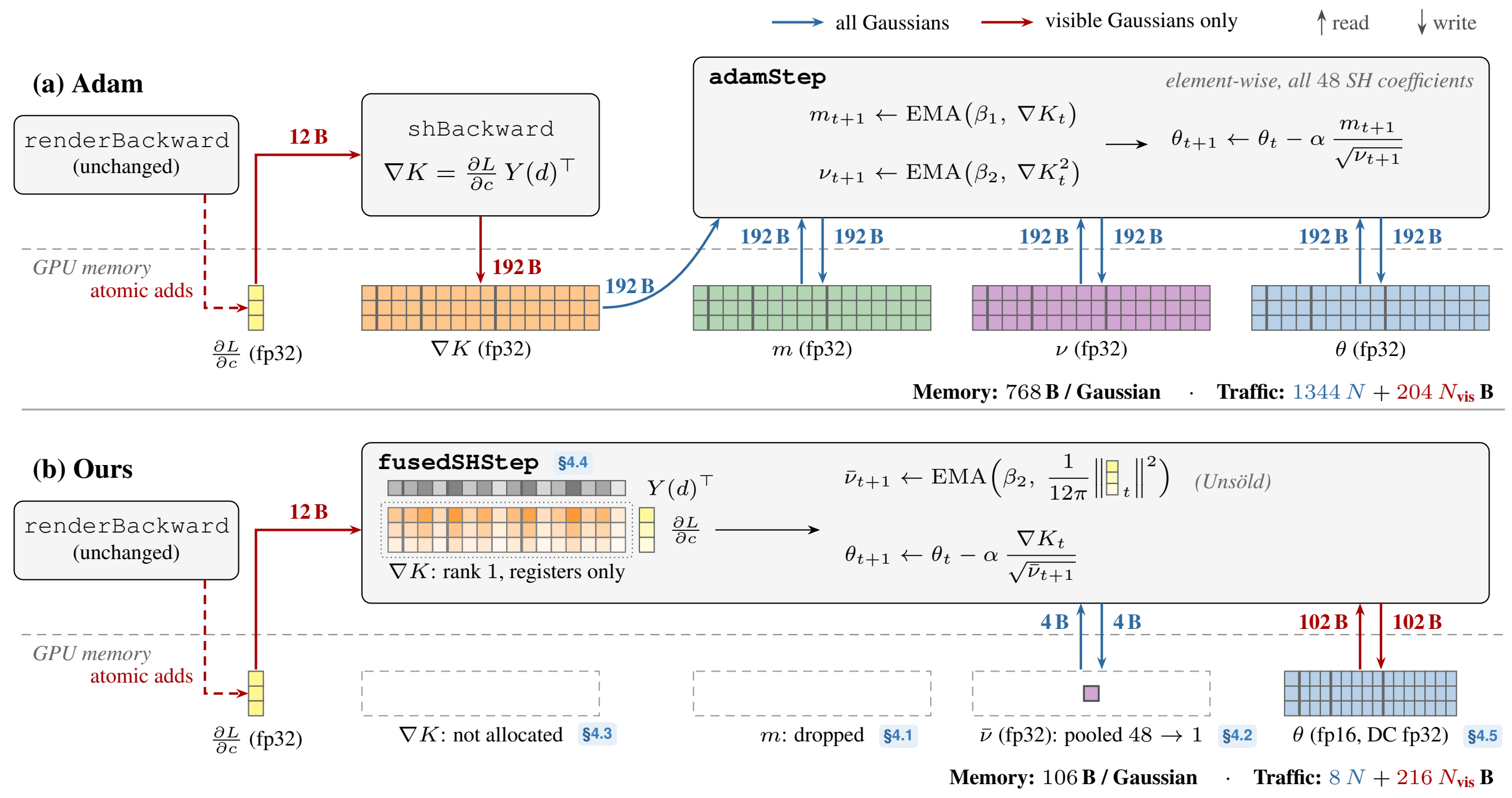
\begin{figure*}[t]
\centering
\usetikzlibrary{arrows.meta,positioning,calc}
\providecolor{linkblue}{rgb}{0.21,0.49,0.74}
\providecommand{\shcell}{0.17}
\providecommand{\methodsec}[2]{\S#2}
\providecommand{\shgrid}[5]{%
  \foreach \c in {0,...,15}{\foreach \r in {0,1,2}{
    \draw[#5,fill=#4] ({#1+\c*\shcell},{#2-\r*#3}) rectangle ++(\shcell,-#3);}}
  \foreach \b in {1,4,9}{
    \draw[#5,line width=0.9pt] ({#1+\b*\shcell},{#2}) -- ++(0,{-3*#3});}
}
\begin{tikzpicture}[
  font=\footnotesize,
  >={Stealth[length=4pt]},
  kernel/.style={draw, rounded corners=3pt, fill=black!4, align=center,
    minimum height=0.9cm, inner sep=3pt},
  mem/.style={densely dashed, black!45},
  lbl/.style={font=\scriptsize, align=center},
  note/.style={font=\scriptsize\itshape, text=black!60, align=center},
  sec/.style={rounded corners=1.5pt, fill=linkblue!12, text=linkblue!80!black,
    font=\sffamily\bfseries\tiny, inner sep=1.5pt},
  all/.style={->, thick, call},
  vis/.style={->, thick, cvis},
  ball/.style={font=\scriptsize\bfseries, text=call},
  bvis/.style={font=\scriptsize\bfseries, text=cvis},
]
\colorlet{cth}{linkblue!35}
\colorlet{cg}{orange!45}
\colorlet{cm}{green!45!black!30}
\colorlet{cv}{violet!35}
\colorlet{cc}{yellow!55}
\colorlet{call}{linkblue!85!black}
\colorlet{cvis}{red!65!black}

\node[anchor=west, font=\small\bfseries] at (-0.7,0.8) {(a) Adam};
\draw[all] (7.9,1.52) -- ++(0.6,0) node[right, lbl, text=black] {all Gaussians};
\draw[vis] (10.3,1.52) -- ++(0.6,0) node[right, lbl, text=black] {visible Gaussians only};
\draw[->, black!70] (14.2,1.37) -- ++(0,0.3) node[pos=0.5, right, lbl] {read};
\draw[->, black!70] (15.35,1.67) -- ++(0,-0.3) node[pos=0.5, right, lbl] {write};
\node[kernel, minimum width=2.2cm] (rbA) at (0.5,0) {\texttt{renderBackward}\\[-1pt]{\scriptsize(unchanged)}};
\node[kernel, minimum width=2.72cm, minimum height=1.4cm] (shA) at (4.56,0) {\texttt{shBackward}\\[5pt]$\nabla K = \frac{\partial L}{\partial c}\,Y(d)^\top$};
\draw[rounded corners=3pt, fill=black!4] (7.0,-0.72) rectangle (16.12,1.12);
\node[anchor=north west, font=\footnotesize\bfseries] at (7.05,1.12) {\texttt{adamStep}};
\node[note, anchor=north east] at (16.07,1.07) {element-wise, all $48$ SH coefficients};
\node[lbl, anchor=west] (mv) at (8.2,0.12) {$\begin{aligned}
  m_{t+1} &\leftarrow \mathrm{EMA}\big(\beta_1,\ \nabla K_t\big)\\[6pt]
  \nu_{t+1} &\leftarrow \mathrm{EMA}\big(\beta_2,\ \nabla K_t^2\big)
\end{aligned}$};
\node[lbl, anchor=west] (theq) at (12.35,0.12) {$\theta_{t+1} \leftarrow \theta_t - \alpha\,\dfrac{m_{t+1}}{\sqrt{\nu_{t+1}}}$};
\draw[->] ([xshift=4pt]mv.east) -- ([xshift=-4pt]theq.west);

\draw[mem] (-0.7,-1.08) -- (16.3,-1.08);
\node[note, anchor=north west] at (-0.7,-1.08) {GPU memory};

\foreach \r in {0,1,2}{\draw[black!60, fill=cc] (1.9,{-1.5-\r*\shcell}) rectangle ++(\shcell,-\shcell);}
\node[lbl, below] at (1.985,-2.01) {$\frac{\partial L}{\partial c}$ (fp32)};
\draw[vis, densely dashed] (1.4,-0.45) |- node[pos=0.42, left, lbl] {atomic adds} (1.9,-1.755);
\draw[vis] (1.985,-1.48) |- node[pos=0.75, above, bvis] {12\,B} (shA.west);
\shgrid{3.2}{-1.5}{\shcell}{cg}{black!60}
\shgrid{7.0}{-1.5}{\shcell}{cm}{black!60}
\shgrid{10.2}{-1.5}{\shcell}{cv}{black!60}
\shgrid{13.4}{-1.5}{\shcell}{cth}{black!60}
\node[lbl, below] at (4.56,-2.01) {$\nabla K$ (fp32)};
\node[lbl, below] at (8.36,-2.01) {$m$ (fp32)};
\node[lbl, below] at (11.56,-2.01) {$\nu$ (fp32)};
\node[lbl, below] at (14.76,-2.01) {$\theta$ (fp32)};

\draw[vis] (shA.south) -- node[pos=0.75, right, bvis] {192\,B} (4.56,-1.48);
\draw[all] (5.96,-1.75) to[out=0, in=-120] node[pos=0.45, left, ball] {192\,B} (7.3,-0.72);
\draw[all] (8.24,-1.48) -- node[pos=0.72, left, ball] {192\,B} (8.24,-0.72);
\draw[all] (8.48,-0.72) -- node[pos=0.28, right, ball] {192\,B} (8.48,-1.48);
\draw[all] (11.44,-1.48) -- node[pos=0.72, left, ball] {192\,B} (11.44,-0.72);
\draw[all] (11.68,-0.72) -- node[pos=0.28, right, ball] {192\,B} (11.68,-1.48);
\draw[all] (14.64,-1.48) -- node[pos=0.72, left, ball] {192\,B} (14.64,-0.72);
\draw[all] (14.88,-0.72) -- node[pos=0.28, right, ball] {192\,B} (14.88,-1.48);
\node[lbl, anchor=north east, align=right, font=\scriptsize\bfseries] at (16.3,-2.5)
  {Memory: $768$\,B / Gaussian \quad$\cdot$\quad Traffic: \textcolor{call}{$1344\,N$} $+$ \textcolor{cvis}{$204\,N_\text{vis}$}\,B};

\draw[black!35, line width=0.6pt] (-0.7,-2.92) -- (16.3,-2.92);
\begin{scope}[yshift=-4.42cm]
\node[anchor=west, font=\small\bfseries] at (-0.7,0.8) {(b) Ours};
\node[kernel, minimum width=2.2cm] (rbB) at (0.5,0) {\texttt{renderBackward}\\[-1pt]{\scriptsize(unchanged)}};

\draw[rounded corners=3pt, fill=black!4] (3.2,-0.72) rectangle (16.12,1.12);
\node[anchor=north west, font=\footnotesize\bfseries] (fk) at (3.25,1.12) {\texttt{fusedSHStep}};
\node[sec, right=1pt of fk] {\methodsec{sec:method_fused_kernel}{4.4}};

\def\ly{0.12}
\def\gx{3.5}
\coordinate (in) at (3.2,\ly);
\foreach \a [count=\r from 0] in {1.0,0.6,0.3}{
  \foreach \b [count=\c from 0] in {0.5,0.55,0.8,0.35,0.9,0.4,0.7,0.25,0.6,0.85,0.3,0.5,0.95,0.45,0.65,0.2}{
    \pgfmathtruncatemacro\p{85*\a*\b}
    \fill[orange!\p!white] ({\gx+\c*\shcell},{\ly+1.5*\shcell-\r*\shcell}) rectangle ++(\shcell,-\shcell);}}
\shgrid{\gx}{\ly+1.5*\shcell}{\shcell}{none}{black!50}
\draw[densely dotted, black!60, rounded corners=1pt] ({\gx-0.08},{\ly+1.5*\shcell+0.06}) rectangle ({\gx+16*\shcell+0.08},{\ly-1.5*\shcell-0.06});
\foreach \b [count=\c from 0] in {0.5,0.55,0.8,0.35,0.9,0.4,0.7,0.25,0.6,0.85,0.3,0.5,0.95,0.45,0.65,0.2}{
  \pgfmathtruncatemacro\p{55*\b}
  \draw[black!60, fill=black!\p] ({\gx+\c*\shcell},{\ly+2.5*\shcell+0.14}) rectangle ++(\shcell,-\shcell);}
\foreach \k in {1,4,9}{\draw[black!60, line width=0.9pt] ({\gx+\k*\shcell},{\ly+2.5*\shcell+0.14}) -- ++(0,-\shcell);}
\foreach \a [count=\r from 0] in {1.0,0.6,0.3}{
  \pgfmathtruncatemacro\p{90*\a}
  \draw[black!60, fill=cc!\p!white] ({\gx+16*\shcell+0.16},{\ly+1.5*\shcell-\r*\shcell}) rectangle ++(\shcell,-\shcell);}
\node[lbl, anchor=west] at ({\gx+16*\shcell+0.12},{\ly+2*\shcell+0.14}) {$Y(d)^\top$};
\node[lbl, anchor=west] at ({\gx+17*\shcell+0.2},\ly) {$\frac{\partial L}{\partial c}$};
\node[lbl, anchor=base] at ({\gx+8*\shcell},{\ly-1.5*\shcell-0.3}) {$\nabla K$: rank $1$, registers only};
\newcommand{\dLdcvec}{\tikz[baseline=-0.6ex]{\foreach \a [count=\r from 0] in {1.0,0.6,0.3}{
  \pgfmathtruncatemacro\p{90*\a}
  \draw[black!60, fill=cc!\p!white] (0,{0.195-\r*0.13}) rectangle ++(0.13,-0.13);}}}
\node[lbl, anchor=west] (upd) at (8.6,0.25) {$\begin{aligned}
  \bar\nu_{t+1} &\leftarrow \mathrm{EMA}\Big(\beta_2,\ \frac{1}{12\pi}\Big\|\dLdcvec_{\,t}\Big\|^2\Big)\\[6pt]
  \theta_{t+1} &\leftarrow \theta_t - \alpha\,\frac{\nabla K_t}{\sqrt{\bar\nu_{t+1}}}
\end{aligned}$};
\node[note, anchor=west] at ([yshift=0.41cm]upd.east) {(Uns\"old)};
\draw[->] (7.25,\ly) -- (7.25+1.2,\ly);

\draw[mem] (-0.7,-1.08) -- (16.3,-1.08);
\node[note, anchor=north west] at (-0.7,-1.08) {GPU memory};

\foreach \r in {0,1,2}{\draw[black!60, fill=cc] (1.9,{-1.5-\r*\shcell}) rectangle ++(\shcell,-\shcell);}
\node[lbl, below] at (1.985,-2.01) {$\frac{\partial L}{\partial c}$ (fp32)};
\draw[vis, densely dashed] (1.4,-0.45) |- node[pos=0.42, left, lbl] {atomic adds} (1.9,-1.755);
\draw[vis] (1.985,-1.48) |- node[pos=0.75, above, bvis] {12\,B} (in);
\draw[densely dashed, black!40] (3.2,-1.5) rectangle ++(2.72,-0.51);
\node[lbl, below] (gl) at (4.56,-2.01) {$\nabla K$: not allocated};
\node[sec, right=1pt of gl] {\methodsec{sec:method_no_grad_memory}{4.3}};
\draw[densely dashed, black!40] (7.0,-1.5) rectangle ++(2.72,-0.51);
\node[lbl, below] (ml) at (8.36,-2.01) {$m$: dropped};
\node[sec, right=1pt of ml] {\methodsec{sec:method_drop_m}{4.1}};
\draw[densely dashed, black!40] (10.2,-1.5) rectangle ++(2.72,-0.51);
\draw[black!70, fill=cv, line width=0.6pt] (11.475,{-1.5-\shcell}) rectangle ++(\shcell,-\shcell);
\node[lbl, below] (vl) at (11.56,-2.01) {$\bar\nu$ (fp32): pooled $48\to1$};
\node[sec, right=1pt of vl] {\methodsec{sec:method_pool_v}{4.2}};
\foreach \r in {0,1,2}{
  \draw[black!60, fill=cth] (13.775,{-1.5-\r*\shcell}) rectangle ++(\shcell,-\shcell);}
\foreach \c in {0,...,14}{\foreach \r in {0,1,2}{
  \draw[black!60, fill=cth] ({13.945+\c*0.12},{-1.5-\r*\shcell}) rectangle ++(0.12,-\shcell);}}
\foreach \b in {0,3,8}{
  \draw[black!60, line width=0.9pt] ({13.945+\b*0.12},-1.5) -- ++(0,{-3*\shcell});}
\node[lbl, below] (tl) at (14.76,-2.01) {$\theta$ (fp16, DC fp32)};
\node[sec, right=1pt of tl] {\methodsec{sec:method_fp16}{4.5}};

\draw[all] (11.44,-1.48) -- node[pos=0.72, left, ball] {4\,B} (11.44,-0.72);
\draw[all] (11.68,-0.72) -- node[pos=0.28, right, ball] {4\,B} (11.68,-1.48);
\draw[vis] (14.64,-1.48) -- node[pos=0.72, left, bvis] {102\,B} (14.64,-0.72);
\draw[vis] (14.88,-0.72) -- node[pos=0.28, right, bvis] {102\,B} (14.88,-1.48);
\node[lbl, anchor=north east, align=right, font=\scriptsize\bfseries] at (16.3,-2.5)
  {Memory: $106$\,B / Gaussian \quad$\cdot$\quad Traffic: \textcolor{call}{$8\,N$} $+$ \textcolor{cvis}{$216\,N_\text{vis}$}\,B};
\end{scope}
\end{tikzpicture}
\caption{Optimizers for the SH coefficients: (a) Adam and (b) Ours.  For
brevity, the geometry parameters are omitted and both methods are shown without
backward fusion (\S\ref{sec:method_fused_kernel}). Kernels are drawn above the
dashed line and per-Gaussian GPU buffers below it; each SH buffer is a
$3\times16$ grid of color channels and SH basis functions, with thick lines
separating the bands. Arrows give the bytes read and written per Gaussian after
\texttt{renderBackward}; the atomic accumulation of $\frac{\partial L}{\partial
c}$ is identical for both and not counted.}
\label{fig:method}
\end{figure*}

We introduce techniques that target the statically allocated optimizer states
and the gradients, rather than the dynamic per-iteration memory
(\S\ref{sec:pre_memory}), to reduce the memory and bandwidth taken by SH. This
section details the techniques we use: dropping the first moment, pooling the
second moment, exploiting the low-rank structure of the SH gradient, and storing
primals in $16$-bits. All other parameters apart from SH
still use the original Adam optimizer.
Figure~\ref{fig:method} gives an overview of our method and shows its effect on
the memory of a Gaussian.  Our method leaves the differentiable rasterization
untouched, and can be used as a drop-in replacement in any large-scale 3DGS
training system that uses Adam to optimize the SH coefficients
(\S\ref{sec:results_other_systems}).

\subsection{Dropping First Moment}
\label{sec:method_drop_m}
The first moment of Adam acts as a low-pass filter on the noisy stochastic
gradients (\S\ref{sec:pre_adam}). We find that for the SH coefficients
this filtering is not necessary and the first moment can be dropped altogether.
The update rule in
Equation~\ref{eq:adam} then reduces to:
$
\theta_{t+1} = \theta_t - \alpha \frac{g_t}{\sqrt{\nu_{t+1}}},
$
which is essentially RMSProp~\cite{tieleman2012rmsprop}. This is not a new
observation---Adafactor~\cite{pmlr-v80-shazeer18a} also drops the first moment
to save memory when training language models.  We show in \S\ref{sec:results}
that it does not hurt the quality of the reconstruction.

A side effect of dropping the first moment is that SH for the invisible
Gaussians outside the view-frustum need not be updated since their gradient is
zero.  Therefore, dropping the first moment, by construction, gives us the
parameter update condition of \textit{Selective Adam} for SH. The second moment
is still updated for all the Gaussians, visible or invisible, since it decays by
$\beta_2$ every iteration (\S\ref{sec:method_fused_kernel}).

\subsection{Pooling Second Moment}
\label{sec:method_pool_v}
The second moment gives each parameter its own effective learning rate by
normalizing the gradient with its running magnitude.  We find that it is enough
to keep a single pooled second moment $\bar\nu$ per Gaussian, maintained as an
EMA of the mean squared gradient over all its $48$ SH coefficients:
\begin{equation}
\boxed{
\begin{aligned}
\bar\nu_{t+1} &= \beta_2 \bar\nu_t + (1 - \beta_2)\, \frac{1}{48}\sum_{i=0}^{47} g_{t,i}^2,\\
\theta_{t+1,i} &= \theta_{t,i} - \alpha \frac{g_{t,i}}{\sqrt{\bar\nu_{t+1}}},
\end{aligned}
}
\label{eq:pool_v}
\end{equation}
where $i$ indexes the SH coefficients of the Gaussian. This reduces the
optimizer state of the SH coefficients from $48$ values to just $1$ per
Gaussian.

Pooling the second moment over a collection of parameters has been explored by
Adam-mini~\cite{zhang2025adammini} for weight matrices in neural networks and by
VectorAdam~\cite{Selena:2022:VFR} to preserve rotation equivariance in vectors.
Our pooling can be viewed as Adam-mini with each Gaussian being treated as its
own independent parameter block. Against VectorAdam, we take a more aggressive
approach by pooling the second moment over all SH basis vectors for all $3$
colors. We also use the mean (instead of sum) to preserve the scale of Adam's
per-parameter second moment estimates.

\subsection{Exploiting Low-Rank SH Gradients}
\label{sec:method_no_grad_memory}
The gradient of the SH coefficients of a Gaussian is the sum of the gradient
contributions of the pixels it splats onto. However, since the SH basis is
evaluated only once for a given camera (\S\ref{sec:pre_rendering}), all
these pixels see the same color $c$ and their gradients accumulate into a single
color gradient $\frac{\partial L}{\partial c} \in \mathbb{R}^3$. As the color is linear in the
SH coefficients (Equation~\ref{eq:sh_color}), the chain rule gives:
\begin{equation}
\dfrac{\partial L}{\partial K} = \dfrac{\partial L}{\partial c}\, Y(d)^\top,
\label{eq:sh_grad_outer}
\end{equation}
i.e., the outer product of the color gradient and the SH basis evaluated at the
viewing direction, which is a rank-$1$ matrix of size $3 \times 16$. Had the SH
been evaluated per pixel, the gradient would instead be a sum of rank-$1$
matrices with a different direction $d_p$ for each pixel.

Existing approaches do not realize this hidden rank-$1$ structure and simply
calculate the full gradient and write it into an $N \times 48$ sized buffer.
Since $Y(d)$ can be evaluated in closed form from the mean of the Gaussian and
the camera center, we only keep the $N \times 3$ color gradient and compute the
gradient of the SH coefficients on-the-fly right when the optimizer actually
requires it. This removes a big bulky statically allocated buffer for SH
coefficients (\S\ref{sec:pre_memory}).

Since we pool the second moment over all the SH coefficients of a Gaussian
(\S\ref{sec:method_pool_v}), this rank-$1$ structure also simplifies its
calculation.  Substituting Equation~\ref{eq:sh_grad_outer} into the pooled
second moment (Equation~\ref{eq:pool_v}), the sum over the $48$ squared entries
of $\frac{\partial L}{\partial K}$ splits into the $3$ color channels and the
$16$ SH basis functions:
\begin{equation}
\frac{1}{48}\sum_{i=0}^{47} g_{t,i}^2 = \frac{1}{48}
\left\|\dfrac{\partial L}{\partial c}\right\|^2 \sum_{j=0}^{15} Y_j(d)^2.
\end{equation}
By Uns\"old's theorem~\cite{unsoldtheorem}, the
squares of the SH basis functions of band $l$ sum up to $(2l+1)/4\pi$ for any
direction, and hence the sum over the $4$ bands is $16/4\pi = 4/\pi$. The pooled
second moment then becomes:
\begin{equation}
\bar\nu_{t+1} = \beta_2 \bar\nu_t + (1 - \beta_2)\, \frac{1}{12\pi}
\left\|\dfrac{\partial L}{\partial c}\right\|^2.
\label{eq:pool_v_fused}
\end{equation}
This replaces the sum over the $16$ squared SH basis functions with a constant,
and $\bar\nu$ only depends on the color gradient of the projected Gaussian,
independent of the viewing direction. This is only possible because of
pooling: with a per-coefficient second moment, each $\nu_i$ sees its own
$Y_j(d)^2$ and there is no sum over the basis functions to collapse.

\subsection{Fused SH Optimizer Kernel}
\label{sec:method_fused_kernel}
Every technique discussed so far---dropping the first moment, pooling the second
moment, and computing the SH gradient on-the-fly---lets us calculate the gradient
and update the SH parameters in a single fused kernel.

The \texttt{renderBackward} kernel performs the alpha-blending yet again and
propagates the gradients from the pixels of the image to the projected 2D
Gaussians~\cite{kerbl3Dgaussians,10.1145/3680528.3687694}. Our kernel takes over
from here. For every Gaussian, it updates the pooled second moment and writes it
back. The invisible Gaussians, which the rasterizer has already marked as
touching no tile, simply decay $\bar\nu$ by $\beta_2$ without reading their
color gradient. For the visible Gaussians, the kernel reads the color gradient,
evaluates $\bar\nu_{t+1}$ from it (Equation~\ref{eq:pool_v_fused}), calculates
the gradient of the SH coefficients on-the-fly by taking the outer product
(Equation~\ref{eq:sh_grad_outer}) and performs the optimizer step
(Equation~\ref{eq:pool_v}), which uses the pooled second moment $\bar\nu_{t+1}$
calculated just before. The total memory bandwidth for this kernel is $8N +
216N_{\text{vis}}$ bytes, while for previous approaches it was $1344N +
204N_{\text{vis}}$ bytes split across multiple kernels. Here, $N$ and
$N_{\text{vis}}$ are the numbers of all and visible Gaussians, respectively.

\textit{Relation to Fused Adam:} Faster-GS~\cite{hahlbohm2026fastergs} keeps
Adam unchanged but updates the visible Gaussians inside the backward pass
(\emph{backward fusion}), so the gradients of the Gaussian parameters, including
the SH gradients, are never stored. Only the gradients of the projected 2D
Gaussians are written, which take $9$ values per Gaussian. However, Adam still
keeps both moments for all $48$ SH coefficients of all $N$ Gaussians, and these
make up most of the \emph{static memory} (\S\ref{sec:pre_memory}). Our fused SH
kernel does not store the SH gradients either, and in addition drops the first
moment and pools the second moment. Backward fusion of the remaining geometry
parameters is independent of this, and we evaluate it for both Adam and ours
(\S\ref{sec:results_standard_benchmark}).

\subsection{Reduced Precision for Primal}
\label{sec:method_fp16}
Finally, we store the $45$ specular SH coefficients in $16$-bit precision to
halve the memory and bandwidth they take. This is a common choice in the
compression works of 3DGS~\cite{3DGSzip2025,PapantonakisReduced3dgs} but
is underexplored in training. We only read and write the primals in $16$-bit,
while all the calculations inside the forward kernel and the fused SH kernel
(\S\ref{sec:method_fused_kernel}) are done in $32$-bit. We do not notice a
quality degradation due to this on common scenes
(\S\ref{sec:results_standard_benchmark}).  However, for large-scale scenes (that
often also require longer training), we see a drop in quality, as the small
late-stage updates fall below the fp16 resolution and get rounded away. We thus
apply stochastic rounding~\cite{3045118.3045303} during writebacks to recover
the quality (see the supplementary).

\section{Validation and Results}
\label{sec:results}

We evaluate our method's (\rom{1}) quality compared to Adam on the standard
datasets, where we stay on par while using $1.7\times$ less memory and
$1.5\times$ less time; (\rom{2}) quality compared to other memory-lean
optimizers, which either match Adam without saving as much or lose quality as
they compress more; (\rom{3}) scaling to more primitives and higher
resolutions, where we fit $105$M Gaussians on a single GPU compared to $40$M
for the prior state-of-the-art and train $1.9$--$2.5\times$ faster; and
(\rom{4}) integration with large scale systems---in CPU-offloaded training, we
reduce PCIe traffic by $3.1$--$3.7\times$ and CPU RAM by $1.7$--$2.6\times$,
and in multi-GPU training, we reduce peak memory per GPU by $1.4$--$1.7\times$.
Our ablation (\S\ref{sec:results_ablation}) shows that each of our techniques
reduces memory and time without regressing quality.

\subsection{Standard Datasets}
\label{sec:results_standard_benchmark}
We benchmark on the three standard 3DGS datasets:
Mip-NeRF 360~\cite{barron2022mipnerf360}, Tanks and
Temples~\cite{Knapitsch2017}, and Deep Blending~\cite{DeepBlending2018}. Beyond
Adam~\cite{kingma2015adam}, we compare against other memory-lean first-order
optimizers used in deep learning: RMSProp~\cite{tieleman2012rmsprop},
Adafactor~\cite{pmlr-v80-shazeer18a}, Adam-mini~\cite{zhang2025adammini}, 8-bit
Adam~\cite{dettmers2022optimizers}, SGD~\cite{Goodfellow-et-al-2016}. We
apply these optimizers only to the $N \times 45$ specular SH
coefficients, since it has its own learning rate. The DC coefficients and
the geometric parameters keep Adam as it is. For Adafactor, we treat the specular
coefficients as an $N \times 45$ matrix; and for Adam-mini, we
partition them into per-Gaussian blocks. PyTorch's
implementation always applies the paper's relative step size, while the
HuggingFace implementation~\cite{wolf-etal-2020-transformers} allows an absolute learning rate. These methods
reduce the optimizer state in a representation agnostic manner whereas our
technique exploits the structure of the SH and its gradient. We also compare
against \textit{Selective Adam} and \textit{ReducedSH Adam}, two 3DGS-specific
optimizers proposed by Taming3DGS~\cite{10.1145/3680528.3687694}. For a fair
comparison of time and memory, we implement every optimizer, as a custom single
kernel for all parameters instead of using their native PyTorch implementation
(which launches several kernels and allocates intermediate buffers). In
Table~\ref{tab:optimizer-comparison}, we report PSNR, SSIM, LPIPS, peak GPU
memory, and total training time. Following Faster-GS, we also test the fused
implementation of the backward and optimizer step for the visible Gaussians in a
training step. All the metrics are averaged over $5$ seeds per scene.  Details
on its fairness are in the supplementary.

The quality does not separate the optimizers by much. Most stay within $0.1$dB
PSNR of Adam. RMSProp and Adam-mini are both on par with Adam; and so is
ours. The optimizers that reduce the storage more aggressively
further lose quality: $8$-bit Adam drops $0.16$dB on T\&T and has worse LPIPS on
all three datasets, and Adafactor drops $0.17$dB on Mip-NeRF 360. However, the
memory usage and training time clearly sets our optimizer apart from others---we
maintain the same quality while using $1.7\times$ less memory and $1.5\times$
less time. The supplementary covers a performance breakdown plot comparing our
method to Fused Adam from Faster-GS.

\providecommand{\optcompfontsize}{\small}
\providecommand{\pl}{\textcolor{gray}{--}}
\providecommand{\fusedrow}{\hspace{0.6em}\textcolor{gray}{$\hookrightarrow$}\,+ bwd.\ fusion}
\begin{table*}[t]
        \centering
        \optcompfontsize
        \setlength{\tabcolsep}{3pt}
        \makebox[\textwidth][c]{\begin{tabular}{l ccccc @{\hskip 8pt} ccccc @{\hskip 8pt} ccccc}
        \toprule
& \multicolumn{5}{c}{Mip-NeRF 360~\cite{barron2022mipnerf360}} & \multicolumn{5}{c}{Tanks \& Temples~\cite{Knapitsch2017}} & \multicolumn{5}{c}{Deep Blending~\cite{DeepBlending2018}} \\
        \cmidrule(lr){2-6} \cmidrule(lr){7-11} \cmidrule(lr){12-16}
Method & PSNR & SSIM & LPIPS & VRAM & Time & PSNR & SSIM & LPIPS & VRAM & Time & PSNR & SSIM & LPIPS & VRAM & Time \\
        \midrule
Adam                   & 29.49 & 0.884 & 0.200 & \cellcolor{tabgrad!19}3.41 & \cellcolor{tabgrad!20}486 & 24.52 & 0.864 & 0.176 & \cellcolor{tabgrad!20}2.02 & \cellcolor{tabgrad!20}303 & 29.79 & 0.905 & 0.300 & \cellcolor{tabgrad!20}3.15 & \cellcolor{tabgrad!22}364 \\
\fusedrow              & 29.49 & 0.884 & 0.200 & \cellcolor{tabgrad!30}2.92 & \cellcolor{tabgrad!23}466 & 24.52 & 0.865 & 0.176 & \cellcolor{tabgrad!34}1.64 & \cellcolor{tabgrad!21}300 & 29.74 & 0.905 & 0.301 & \cellcolor{tabgrad!33}2.60 & \cellcolor{tabgrad!31}318 \\
RMSProp      & 29.51 & 0.884 & 0.200 & \cellcolor{tabgrad!29}2.98 & \cellcolor{tabgrad!26}448 & 24.54 & 0.865 & 0.175 & \cellcolor{tabgrad!29}1.76 & \cellcolor{tabgrad!26}278 & 29.83 & 0.905 & 0.301 & \cellcolor{tabgrad!30}2.72 & \cellcolor{tabgrad!30}323 \\
Adam-mini    & 29.46 & 0.884 & 0.200 & \cellcolor{tabgrad!29}2.98 & \cellcolor{tabgrad!26}448 & 24.48 & 0.864 & 0.177 & \cellcolor{tabgrad!29}1.77 & \cellcolor{tabgrad!26}278 & 29.79 & 0.906 & 0.300 & \cellcolor{tabgrad!30}2.73 & \cellcolor{tabgrad!30}322 \\
Adafactor    & 29.32 & 0.883 & 0.201 & \cellcolor{tabgrad!38}2.57 & \cellcolor{tabgrad!10}549 & 24.45 & 0.863 & 0.178 & \cellcolor{tabgrad!39}1.51 & \cellcolor{tabgrad!10}345 & 29.64 & 0.905 & 0.302 & \cellcolor{tabgrad!40}2.32 & \cellcolor{tabgrad!10}422 \\
Adafactor (HF)& 29.51 & 0.883 & 0.200 & \cellcolor{tabgrad!38}2.56 & \cellcolor{tabgrad!21}481 & 24.62 & 0.865 & 0.174 & \cellcolor{tabgrad!39}1.51 & \cellcolor{tabgrad!21}299 & 29.70 & 0.905 & 0.301 & \cellcolor{tabgrad!40}2.31 & \cellcolor{tabgrad!24}354 \\
8-bit Adam   & 29.51 & 0.880 & 0.207 & \cellcolor{tabgrad!36}2.65 & \cellcolor{tabgrad!27}439 & 24.36 & 0.859 & 0.186 & \cellcolor{tabgrad!37}1.55 & \cellcolor{tabgrad!29}267 & 29.92 & 0.907 & 0.308 & \cellcolor{tabgrad!35}2.54 & \cellcolor{tabgrad!36}295 \\
SGD          & 29.28 & 0.882 & 0.202 & \cellcolor{tabgrad!40}2.50 & \cellcolor{tabgrad!32}409 & 24.52 & 0.864 & 0.176 & \cellcolor{tabgrad!39}1.51 & \cellcolor{tabgrad!32}257 & 29.74 & 0.906 & 0.301 & \cellcolor{tabgrad!41}2.28 & \cellcolor{tabgrad!38}282 \\
Selective Adam & 29.27 & 0.875 & 0.213 & \cellcolor{tabgrad!23}3.25 & \cellcolor{tabgrad!43}336 & 24.29 & 0.856 & 0.191 & \cellcolor{tabgrad!24}1.91 & \cellcolor{tabgrad!43}211 & 29.86 & 0.904 & 0.308 & \cellcolor{tabgrad!18}3.26 & \cellcolor{tabgrad!46}245 \\
ReducedSH              & 29.47 & 0.884 & 0.200 & \cellcolor{tabgrad!10}3.84 & \cellcolor{tabgrad!30}417 & 24.37 & 0.863 & 0.178 & \cellcolor{tabgrad!10}2.28 & \cellcolor{tabgrad!31}260 & 29.90 & 0.905 & 0.299 & \cellcolor{tabgrad!10}3.58 & \cellcolor{tabgrad!35}296 \\
        \midrule
Ours         & 29.46 & 0.884 & 0.201 & \cellcolor{tabgrad!55}1.82 & \cellcolor{tabgrad!44}330 & 24.52 & 0.864 & 0.176 & \cellcolor{tabgrad!55}1.08 & \cellcolor{tabgrad!46}197 & 29.90 & 0.906 & 0.302 & \cellcolor{tabgrad!57}1.59 & \cellcolor{tabgrad!52}211 \\
\fusedrow              & 29.47 & 0.884 & 0.201 & \cellcolor{tabgrad!57}1.71 & \cellcolor{tabgrad!47}308 & 24.50 & 0.864 & 0.176 & \cellcolor{tabgrad!58}1.00 & \cellcolor{tabgrad!49}185 & 29.76 & 0.905 & 0.303 & \cellcolor{tabgrad!60}1.47 & \cellcolor{tabgrad!56}193 \\
        \bottomrule
        \end{tabular}}%
        \caption{We compare different optimizers for SH on the
        standard datasets, averaged over $5$ seeds, with MCMC densification.
        \protect$\hookrightarrow$\,\textit{+ bwd.\
        fusion} fuses the gradient backprop and optimizer update of all
        parameters of the visible Gaussians~\cite{hahlbohm2026fastergs}.
        \textit{VRAM}: peak GPU memory (GB); \textit{Time}: total training time
        (s). Each optimizer either uses the learning rate proposed for 3DGS
        or a grid-searched one.}
        \label{tab:optimizer-comparison}
        \end{table*}

\subsection{Scaling Up Behaviour}
\label{sec:results_scaling}
We test our optimizer with backward fusion against the fused backward
Adam of Faster-GS~\cite{hahlbohm2026fastergs}. We sweep the converged primitive
count at $3$ fixed resolutions, and report peak memory and training time for
both optimizers in Fig.~\ref{fig:res_prim_sweep}.  We run this experiment on
\textit{bicycle} at the \textit{quarter} resolution ($1237 \times 822$)---the
standard Mip-NeRF 360 protocol.  We also study the behaviour at higher
resolutions: \textit{half} ($2473 \times 1643$) and \textit{full} ($4946 \times
3286$). For every resolution and Gaussian count, we re-train a \emph{separate}
model on a single L40S for $30$k iterations, increasing the Gaussian count in
steps of $5$M until the GPU runs out of memory.  We design this experiment to
test the limits of the system from a performance perspective and not the
quality, hence it uses a modified densification schedule (see the supplementary
for more details) to reach the target Gaussian count. At the standard
\textit{quarter} resolution, the existing state-of-the-art fits at most $40$M
Gaussians, while our optimizer reaches $105$M on the same chip and trains
$1.9$--$2.5\times$ faster.  At $40$M Gaussians, our peak memory is $2.5\times$
lower ($16.4$ vs.\ $41.7$GB). At \textit{half} and \textit{full} resolution, we
fit $2.5\times$ and $2.0\times$ more Gaussians and train $1.5$--$2.1\times$ and
$1.0$--$1.7\times$ faster. The speedup is smaller at higher resolution, since
the differentiable rasterization takes up a larger share of the iteration time.
Additional details and discussion are in the supplementary.

\definecolor{plotfgf}{HTML}{0072B2}
\definecolor{plotours}{HTML}{D55E00}
\begin{figure}[t]
\centering
\ifdefined\sweeplegend\else\newsavebox\sweeplegend\fi
\newcommand*{\sweepsample}[5]{%
  \tikz[baseline=-0.6ex]\draw[#1, line width=#2, draw opacity=0.5, #3]
    plot[mark=#4, mark size=#5, mark indices={2}, mark options={solid, fill=#1,
    draw=white, line width=0.25pt, draw opacity=1, fill opacity=1}]
    coordinates {(0,0) (0.28,0) (0.56,0)};}
\newcommand*{\sweepfgf}[1]{\sweepsample{plotfgf}{0.9pt}{#1}{triangle*}{1.8pt}}
\newcommand*{\sweepours}[1]{\sweepsample{plotours}{1.4pt}{#1}{square*}{1.3pt}}
\sbox\sweeplegend{\scriptsize\setlength{\tabcolsep}{3pt}%
  \renewcommand{\arraystretch}{0.9}%
  \begin{tabular}{@{}lccc@{}}
             & $1.02$\,MP       & $4.06$\,MP         & $16.25$\,MP \\
  \textsc{{[}Adam + bwd.\ fusion{]}} & \sweepfgf{solid}  & \sweepfgf{dashed}  & \sweepfgf{densely dotted} \\
  \textsc{{[}Ours + bwd.\ fusion{]}} & \sweepours{solid} & \sweepours{dashed} & \sweepours{densely dotted} \\
  \end{tabular}}
\begin{tikzpicture}
\tikzset{
  gpurule/.style={gray!55, dashed, line width=0.4pt},
  oom/.style={only marks, mark=x, mark size=2.6pt, line width=0.9pt},
  paneltitle/.style={font=\footnotesize, anchor=north west, inner sep=1.5pt,
    fill=white, at={(rel axis cs:0,1)}, xshift=1.5pt, yshift=-1.5pt},
}
\begin{groupplot}[
  group style={group size=1 by 2, vertical sep=0.25cm,
    xlabels at=edge bottom, xticklabels at=edge bottom},
  scale only axis=true, width=\columnwidth-24pt,
  xmin=0, xmax=113, xtick={0,20,40,60,80,100},
  xmajorgrids=true,
  grid style={gray!20, line width=0.3pt},
  axis line style={gray!60, line width=0.4pt},
  tick label style={font=\scriptsize}, label style={font=\footnotesize},
  every axis plot/.append style={mark=none},
  fgf/.style={plotfgf, line width=0.9pt, draw opacity=0.5,
    mark=triangle*, mark size=1.8pt, mark options={solid, fill=plotfgf, draw=white,
    line width=0.25pt, draw opacity=1, fill opacity=1}},
  ours/.style={plotours, line width=1.4pt, draw opacity=0.5,
    mark=square*, mark size=1.3pt, mark options={solid, fill=plotours,
    draw=white, line width=0.25pt, draw opacity=1, fill opacity=1}},
  oomseg/.style={mark=none, forget plot},
  half/.style={dashed},
  speedup/.style={black!75, line width=1.1pt, draw opacity=0.5,
    mark=*, mark size=1.4pt, mark options={solid, fill=black!75, draw=white,
    line width=0.25pt, draw opacity=1, fill opacity=1}},
  quarter/.style={solid},
  full/.style={densely dotted},
]

\nextgroupplot[height=3.6cm,
  ymin=0, ymax=60, ytick={0,12,24,44.4}, yticklabels={0,12,24,44.4},
]
\node[paneltitle] {Peak VRAM (GB)};
\addplot[gpurule, forget plot] coordinates {(0,12) (113,12)};
\addplot[gpurule, forget plot] coordinates {(0,24) (113,24)};
\addplot[gray!70, line width=0.6pt, forget plot] coordinates {(0,44.4) (113,44.4)};
\node[font=\tiny, text=gray!80!black, anchor=south east, inner sep=1pt]
  at (axis cs:113,12) {12\,GB card};
\node[font=\tiny, text=gray!80!black, anchor=south east, inner sep=1pt]
  at (axis cs:113,24) {24\,GB card};

\addplot[fgf, full] coordinates {(5,8) (10,12.9) (15,17.7) (20,22.3) (25,27) (30,31.6) (35,36.5) (40,41.3)};
\addplot[fgf, full, oomseg] coordinates {(40,41.3) (45,44.4)};
\addplot[ours, full] coordinates {(5,5.8) (10,8.6) (15,11.2) (20,13.7) (25,16.2) (30,18.4) (35,21.4) (40,23.9) (45,26.3) (50,28.7) (55,31) (60,33.6) (65,36.2) (70,38.4) (75,41) (80,43.3)};
\addplot[ours, full, oomseg] coordinates {(80,43.3) (85,44.4)};
\addplot[fgf, half] coordinates {(5,5.2) (10,10.4) (15,15.5) (20,20.7) (25,25.9) (30,31.1) (35,36.2) (40,41.6)};
\addplot[fgf, half, oomseg] coordinates {(40,41.6) (45,44.4)};
\addplot[ours, half] coordinates {(5,3) (10,5.3) (15,7.5) (20,9.7) (25,11.8) (30,13.9) (35,16.1) (40,18.2) (45,20.3) (50,22.3) (55,24.4) (60,26.5) (65,28.6) (70,30.7) (75,32.9) (80,34.9) (85,37) (90,39.1) (95,40.8) (100,43.2)};
\addplot[ours, half, oomseg] coordinates {(100,43.2) (105,44.4)};
\addplot[fgf, quarter] coordinates {(5,5.2) (10,10.4) (15,15.6) (20,20.8) (25,26) (30,31.2) (35,36.5) (40,41.7)};
\addplot[fgf, quarter, oomseg] coordinates {(40,41.7) (45,44.4)};
\addplot[ours, quarter] coordinates {(5,2.3) (10,4.3) (15,6.4) (20,8.3) (25,10.3) (30,12.2) (35,14.3) (40,16.4) (45,18.4) (50,20.5) (55,22.6) (60,24.6) (65,26.8) (70,28.8) (75,30.9) (80,33) (85,35.1) (90,37.2) (95,39.3) (100,41.5) (105,43.5)};
\addplot[ours, quarter, oomseg] coordinates {(105,43.5) (110,44.4)};

\addplot[oom, plotfgf, forget plot] coordinates {(45,44.4)};
\addplot[oom, plotours, forget plot] coordinates {(85,44.4) (105,44.4) (110,44.4)};

\draw[<->, >=stealth, gray!80!black, line width=0.5pt, densely dotted]
  (axis cs:40.8,47.5) -- node[above, font=\scriptsize, text=black, inner sep=1pt]
  {$2.0\times$ Gaussians} (axis cs:79.2,47.5);
\draw[<->, >=stealth, gray!80!black, line width=0.5pt]
  (axis cs:40.8,54) -- node[above, font=\scriptsize, text=black, inner sep=1pt]
  {$2.6\times$ Gaussians} (axis cs:104.2,54);

\nextgroupplot[height=2.0cm, xlabel={Gaussians (M)},
  ymin=0.9, ymax=2.7, ytick={1,1.5,2,2.5}, yticklabels={$1\times$,$1.5\times$,$2\times$,$2.5\times$},
  ymajorgrids=true]
\addplot[speedup, full] coordinates {(5,1.04) (10,1.22) (15,1.32) (20,1.36) (25,1.48) (30,1.52) (35,1.55) (40,1.65)};
\addplot[speedup, half] coordinates {(5,1.50) (10,1.72) (15,1.87) (20,1.95) (25,1.99) (30,2.06) (35,2.12) (40,2.13)};
\addplot[speedup, quarter] coordinates {(5,1.90) (10,2.16) (15,2.27) (20,2.33) (25,2.39) (30,2.44) (35,2.45) (40,2.51)};
\node[font=\tiny, text=black!75, anchor=west, inner sep=1pt] at (axis cs:40.5,2.51) {$1.02$\,MP};
\node[font=\tiny, text=black!75, anchor=west, inner sep=1pt] at (axis cs:40.5,2.13) {$4.06$\,MP};
\node[font=\tiny, text=black!75, anchor=west, inner sep=1pt] at (axis cs:40.5,1.65) {$16.25$\,MP};
\node[font=\scriptsize\itshape, text=gray!80!black, anchor=west]
  at (axis cs:45,1.2) {\textup{\textsc{[Adam + bwd.\ fusion]}} out of memory};
\node[paneltitle] {Speedup};
\end{groupplot}
\node[anchor=south, inner sep=0pt] at ([yshift=3pt]group c1r1.north) {\usebox\sweeplegend};
\end{tikzpicture}
\caption{Peak memory and training time speedup of \textsc{[Ours + bwd.\ fusion]} over
\textsc{[Adam + bwd.\ fusion]} against the Gaussian count on
\textit{bicycle} at three resolutions, until the L40S chip goes OOM.  Each
marker is a separate training run. At the quarter resolution, our
optimizer fits $2.6\times$ more Gaussians and trains $1.9$--$2.5\times$ faster,
and only improves with scale.
}
\label{fig:res_prim_sweep}
\end{figure}
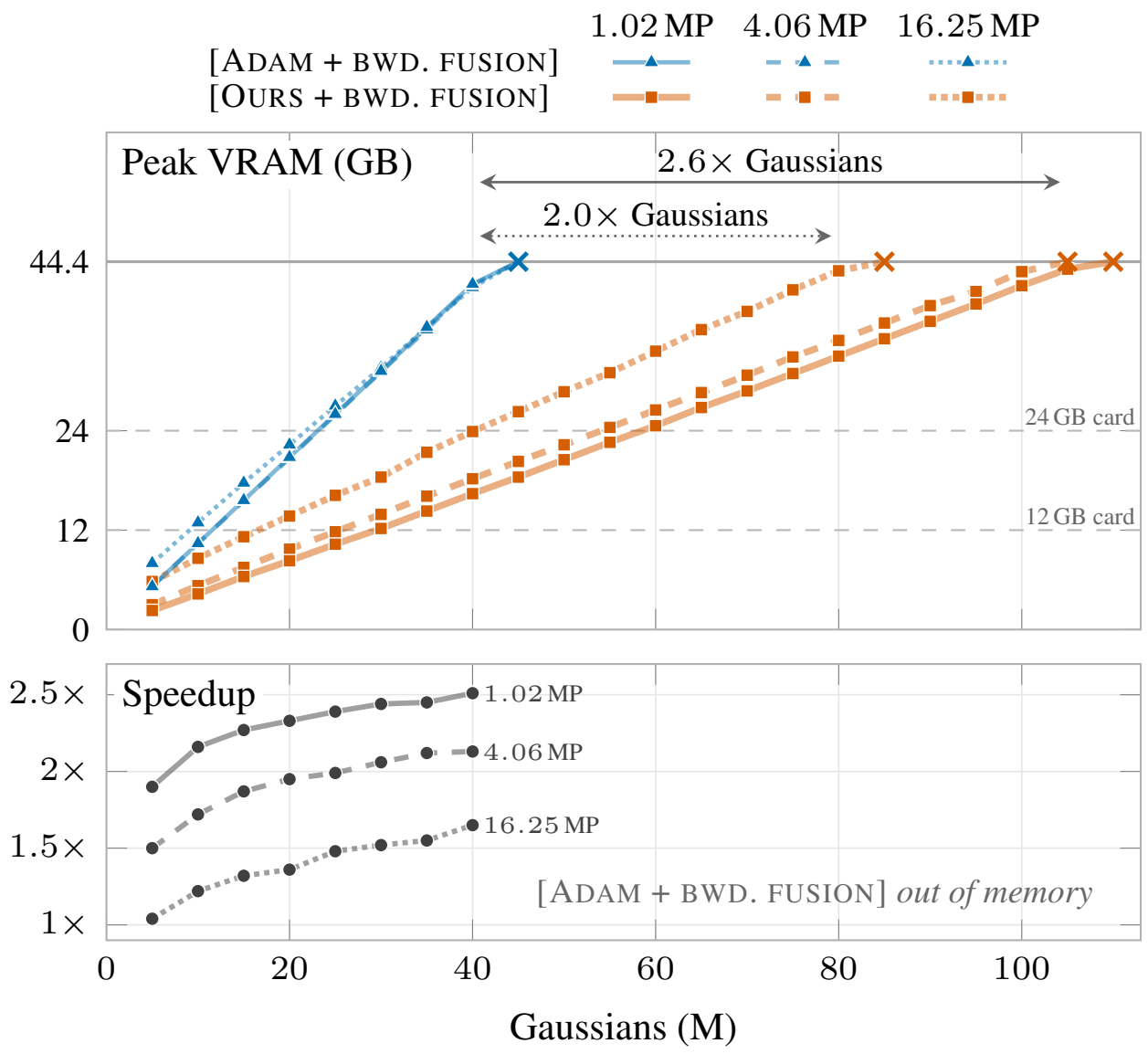

\subsection{Integration with Other Training Systems}
\label{sec:results_other_systems}
Our system is built on top of Faster-GS~\cite{hahlbohm2026fastergs} which is
hyper-engineered for a single GPU. Since the differentiable rasterization is
fast, the optimizer (Adam) is the bottleneck. Other training systems for large scenes spend more time elsewhere---host-device transfers, CPU-side
scheduling or inter-GPU communication---and the end-to-end speedup from a faster
optimizer is bounded accordingly. However, the memory savings do not depend on
this, since SH-related memory is still $4 \times N \times 48 \times 4$ bytes
in these systems. We integrate our method into two such systems and report
memory and communication savings and speedup.

\textit{CPU-Offloaded Training:} CLM~\cite{zhao2025clm} offloads the color
coefficients to the host and the SH gradients are transferred over PCIe
to the CPU for the optimizer using Selective Adam. With the low-rank
structure (\S\ref{sec:method_no_grad_memory}), we instead transfer only the color gradient of the projected Gaussian $\frac{\partial L}{\partial c} \in \mathbb{R}^3$ for each primitive,
instead of the $48$ SH gradients---$16\times$ reduction in the backward
traffic for color. We store the pooled optimizer states on the CPU and the
fused SH kernel (\S\ref{sec:method_fused_kernel}) runs there, computing the SH
gradient on-the-fly, i.e., no memory for gradients is allocated on the CPU
RAM. The $16$-bit SH primals reduce the PCIe transfer for the forward pass.
See the supplementary for details. Table~\ref{tab:clmgs} shows on-par quality on MatrixCity (big city, aerial views)~\cite{li2023matrixcity}, with $3.1$--$3.7\times$ less PCIe traffic since we transfer only $\frac{\partial L}{\partial c}$. The peak CPU RAM reduces $1.7$--$2.6\times$ due to shrinking of SH optimizer states and gradient. The original CLM implementation does
not flush subnormal values to zero, which slows down its training by up to
$1.4\times$; we fix this (Sel.\ Adam$^\dagger$) for a fair comparison. Against
this baseline we train up to $1.5\times$ faster, with the gap growing with the
primitive count.

\begin{table}[t]
  \centering
  \small
  \setlength{\tabcolsep}{1.9pt}
  \begin{tabular}{@{}c|l|ccc|cccc@{}}
  \hline
   & & PSNR & SSIM & LPIPS & PCIe & RAM & VRAM & Time$^*$ \\
  \hline
   & Sel.\ Adam & 25.15 & 0.719 & 0.478 & \cellcolor{tabgrad!10}34.6 & \cellcolor{tabgrad!10}35.4 & \cellcolor{tabgrad!10}7.0 & \cellcolor{tabgrad!10}131 \\
   & Sel.\ Adam$^\dagger$ & 25.19 & 0.719 & 0.478 & \cellcolor{tabgrad!12}33.6 & \cellcolor{tabgrad!15}32.6 & \cellcolor{tabgrad!10}7.0 & \cellcolor{tabgrad!16}119 \\
  \multirow{-3}{*}{\rotatebox[origin=c]{90}{$25$M}}
   & Ours & 25.15 & 0.718 & 0.478 & \cellcolor{tabgrad!56}11.0 & \cellcolor{tabgrad!40}19.7 & \cellcolor{tabgrad!10}7.0 & \cellcolor{tabgrad!18}116 \\
  \hline
   & Sel.\ Adam & 25.56 & 0.742 & 0.450 & \cellcolor{tabgrad!10}65.9 & \cellcolor{tabgrad!13}48.7 & \cellcolor{tabgrad!10}13.1 & \cellcolor{tabgrad!10}265 \\
   & Sel.\ Adam$^\dagger$ & 25.57 & 0.742 & 0.450 & \cellcolor{tabgrad!13}62.8 & \cellcolor{tabgrad!10}51.0 & \cellcolor{tabgrad!11}13.0 & \cellcolor{tabgrad!28}193 \\
  \multirow{-3}{*}{\rotatebox[origin=c]{90}{$51$M}}
   & Ours & 25.57 & 0.742 & 0.450 & \cellcolor{tabgrad!59}18.0 & \cellcolor{tabgrad!45}24.5 & \cellcolor{tabgrad!12}12.7 & \cellcolor{tabgrad!38}154 \\
  \hline
   & Sel.\ Adam & 25.89 & 0.761 & 0.426 & \cellcolor{tabgrad!10}127.1 & \cellcolor{tabgrad!10}89.0 & \cellcolor{tabgrad!10}25.1 & \cellcolor{tabgrad!10}461 \\
   & Sel.\ Adam$^\dagger$ & 25.95 & 0.761 & 0.426 & \cellcolor{tabgrad!13}121.9 & \cellcolor{tabgrad!12}86.9 & \cellcolor{tabgrad!10}25.1 & \cellcolor{tabgrad!28}336 \\
  \multirow{-3}{*}{\rotatebox[origin=c]{90}{$102$M}}
   & Ours & 25.96 & 0.760 & 0.427 & \cellcolor{tabgrad!60}32.7 & \cellcolor{tabgrad!53}32.8 & \cellcolor{tabgrad!11}24.6 & \cellcolor{tabgrad!44}225 \\
  \hline
  \end{tabular}
  \caption{In CLM~\cite{zhao2025clm}, our optimizer matches the quality of
  Selective Adam with heavy reduction in PCIe traffic, host RAM, and training
  time scaling with $3$ increasing budgets on MatrixCity
  (big city, aerial views)~\cite{li2023matrixcity}.  Sel.\ Adam$^\dagger$ is Selective Adam
  running on CPU with subnormal values flushed to zero.  \textit{PCIe}: total
  host--device traffic (TB); \textit{RAM}/\textit{VRAM}: peak host/GPU memory
  (GB); \textit{Time*}: total training time (min); \textit{$^*$}noisy due to
  PCIe contention with potential other jobs on a cluster node.}
  \label{tab:clmgs}
\end{table}

\begin{table}[t]
  \centering
  \small
  \setlength{\tabcolsep}{3pt}
  \begin{tabular}{@{}cl|ccc|cc@{}}
  \hline
   & & PSNR & SSIM & LPIPS & VRAM & Time \\
  \hline
   & Adam & 27.32 & 0.821 & 0.300 & \cellcolor{tabgrad!10}11.0 & \cellcolor{tabgrad!10}8.8 \\
  \multirow{-2}{*}{\shortstack{Rubble4k\\{\footnotesize $28.0$M}}}
   & Ours & 27.27 & 0.821 & 0.300 & \cellcolor{tabgrad!60}6.3 & \cellcolor{tabgrad!46}6.1 \\
  \hline
   & Adam & 22.39 & 0.746 & 0.308 & \cellcolor{tabgrad!10}15.3 & \cellcolor{tabgrad!10}14.5 \\
  \multirow{-2}{*}{\shortstack{Uni10k\\{\footnotesize $28.6$M}}}
   & Ours & 22.36 & 0.745 & 0.308 & \cellcolor{tabgrad!43}11.0 & \cellcolor{tabgrad!30}12.0 \\
  \hline
   & Adam & 26.74 & 0.837 & 0.228 & \cellcolor{tabgrad!10}17.3 & \cellcolor{tabgrad!10}1.9 \\
  \multirow{-2}{*}{\shortstack{MatrixCity\\{\footnotesize $36.6$M}}}
   & Ours & 26.76 & 0.839 & 0.226 & \cellcolor{tabgrad!47}11.9 & \cellcolor{tabgrad!28}1.6 \\
  \hline
  \end{tabular}
  \caption{In Grendel, our optimizer matches Adam's quality while reducing memory and time. We evaluate $3$ large-scale scenes and achieve on-par quality. These runs use modified Grendel code running distributed training on $4$ RTX PRO 6000
  Blackwell GPUs. \textit{VRAM}: peak GPU memory per rank (GB); \textit{Time}: training time (GPU-hours).}
  \label{tab:grendel}
\end{table}

\providecommand{\pl}{\textcolor{gray}{--}}
\begin{table}[t]
        \centering
        \small
        \setlength{\tabcolsep}{3pt}
        \makebox[\columnwidth][c]{\begin{tabular}{l c ccccc}
        \toprule
Variant & SH & PSNR & SSIM & LPIPS & VRAM & Time \\
        \midrule
Adam                   & 768 & 28.64 & 0.884 & 0.214 & \cellcolor{tabgrad!10}3.11 & \cellcolor{tabgrad!10}431 \\
+ drop $m$             & 576 & 28.68 & 0.885 & 0.214 & \cellcolor{tabgrad!24}2.67 & \cellcolor{tabgrad!18}396 \\
+ pool $\nu$           & 388 & 28.60 & 0.884 & 0.215 & \cellcolor{tabgrad!37}2.26 & \cellcolor{tabgrad!26}360 \\
+ fused SH kernel      & 196 & 28.61 & 0.884 & 0.215 & \cellcolor{tabgrad!50}1.86 & \cellcolor{tabgrad!42}292 \\
+ 16-bit primal        & 106 & 28.64 & 0.884 & 0.215 & \cellcolor{tabgrad!57}1.64 & \cellcolor{tabgrad!44}284 \\
+ bwd.\ fusion         & 106 & 28.62 & 0.884 & 0.215 & \cellcolor{tabgrad!60}1.54 & \cellcolor{tabgrad!48}264 \\
        \bottomrule
        \end{tabular}}%
        \caption{Ablation study averaged over $11$ scenes
        and $5$ seeds. Each row adds a new component to previous one,
        starting with Adam. The last row adds backward fusion of remaining
        geometry parameters from Faster-GS~\cite{hahlbohm2026fastergs}; the
        fused SH kernel runs in SH backward pass. \textit{SH}:
        bytes/Gaussian for SH primals, gradients and optimizer states;
        \textit{VRAM}: peak memory (GB); \textit{Time}: training time
        (s).}
        \label{tab:ablation}
        \end{table}

\textit{Multi-GPU Training:} Grendel~\cite{grendelgs} shards the Gaussians
and their optimizer states across GPUs (\S\ref{sec:rw_large_scenes}). We
first port Faster-GS's optimizations into Grendel without changing the math (see
the supplementary). We then measure the change due
to our optimizer on Rubble4k, Uni10k and MatrixCity (Table~\ref{tab:grendel}).
We maintain the quality and reduce the peak GPU memory per rank by $1.7\times$
on Rubble4k, $1.4\times$ on Uni10k and $1.5\times$ on MatrixCity with fewer
GPU-hours.

\subsection{Ablation Study}
\label{sec:results_ablation}

We ablate the techniques in \S\ref{sec:method} by adding them one at
a time to Adam for the SH coefficients, until we arrive
at our full optimizer. Table~\ref{tab:ablation} reports the statically allocated
bytes per Gaussian, reconstruction quality, peak memory, and training time as we
step through our proposed techniques. Results are averaged over the
$11$ scenes from the 3 standard datasets in
\S\ref{sec:results_standard_benchmark}, each with $5$ seeds.
Since the fused SH kernel already runs in the backward pass, the last row only
adds backward fusion of geometry and the SH bytes remain the same, while the
gradients of the means, scales and rotations are no longer stored. All
techniques reduce the static memory allocation, the peak memory utilization and
training time, without regressing quality.

\section{Conclusion}
\label{sec:conclusion}

We presented SHrank---a memory-efficient optimizer for Spherical
Harmonics in 3DGS. Our key observation is that SH coefficients need neither
momentum nor per-coefficient adaptivity, and their gradient is rank-$1$ by
construction, eliminating gradient storage. When coupled with Uns\"old's theorem
and $16$-bit primals, it collapses the SH optimizer into a single fused kernel
that matches Adam's quality at a fraction of its memory and bandwidth. Since
SHrank keeps the representation unchanged, it composes with other systems and
makes large-scale reconstruction more accessible.

\paragraph{Limitations.}
The rank-$1$ gradient, and hence Uns\"old's shortcut and the fused kernel, rely
on SH being evaluated once per view with a single direction $d$ per Gaussian
(\S\ref{sec:pre_rendering}). Per-pixel color evaluation and other
appearance models (\eg, Spherical Gaussians, neural features) lose this
structure. The end-to-end speedup is smaller in systems where the optimizer is
not the bottleneck. Finally, our method is justified only empirically, and a
theoretical analysis of why the SH coefficients need neither momentum nor
per-coefficient adaptivity is left for future work.

\section*{Acknowledgments}
Rahul Goel was funded by the Defense Advanced Research Projects Agency (DARPA)
under the MOCHA program (HR001125CE038) and the National Science Foundation
under grant OAC2403239. Ahmed H. Mahmoud was funded by the National Science
Foundation under grant OAC2403239 and the MIT Generative AI Impact Consortium.

{
    \small
    \bibliographystyle{ieeenat_fullname}
    \bibliography{references}
}

\clearpage
\setcounter{section}{0}
\setcounter{figure}{0}
\setcounter{table}{0}
\setcounter{equation}{0}
\renewcommand{\thesection}{S\arabic{section}}
\renewcommand{\thefigure}{S\arabic{figure}}
\renewcommand{\thetable}{S\arabic{table}}
\renewcommand{\theequation}{S\arabic{equation}}
\renewcommand{\theHsection}{S\arabic{section}}
\renewcommand{\theHfigure}{S\arabic{figure}}
\renewcommand{\theHtable}{S\arabic{table}}
\renewcommand{\theHequation}{S\arabic{equation}}

\maketitlesupplementary

\section{Experimental Details}
\label{sec:supp_experimental_details}
Here, we describe the setup shared by all experiments unless stated otherwise.
Table~\ref{tab:supp-datasets} summarizes the datasets used.

\begin{table}[t]
  \centering
  \small
  \setlength{\tabcolsep}{3pt}
  \begin{tabular}{@{}lcc@{}}
  \toprule
  Dataset & \#Images & MP \\
  \midrule
  Mip-NeRF 360 (outdoor, $1/4$)~\cite{barron2022mipnerf360} & $168$ & $1.04$ \\
  Mip-NeRF 360 (indoor, $1/2$)~\cite{barron2022mipnerf360}  & $281$ & $1.62$ \\
  Tanks \& Temples~\cite{Knapitsch2017}               & $276$ & $0.53$ \\
  Deep Blending~\cite{DeepBlending2018}                & $244$ & $1.11$ \\
  \midrule
  MatrixCity (small, aerial)~\cite{li2023matrixcity} & $6{,}362$ & $2.07$ \\
  MatrixCity (big, aerial)~\cite{li2023matrixcity}   & $60{,}698$ & $2.07$ \\
  Rubble4k~\cite{meganerf}                                & $1{,}678$ & $15.77$ \\
  Uni10k~\cite{windisch2026ALoDOfGaussians}               & $10{,}063$ & $8.18$ \\
  \bottomrule
  \end{tabular}
  \caption{Number of images and image resolution (in megapixels, MP) of each dataset. For Mip-NeRF 360, Tanks \& Temples and Deep Blending, both are averaged over the scenes.}
  \label{tab:supp-datasets}
\end{table}

\textit{Standard Datasets:} We use the $7$ public scenes of Mip-NeRF
360~\cite{barron2022mipnerf360} (bicycle, garden, stump, room, counter, kitchen,
bonsai), $2$ scenes of Tanks \& Temples~\cite{Knapitsch2017} (train and truck)
and the $2$ scenes of Deep Blending~\cite{DeepBlending2018} (drjohnson and
playroom)---resulting in $11$ scenes in total. Following
3DGS~\cite{kerbl3Dgaussians}, we hold out every $8$th image of each scene for
testing and we use the remaining images for training.

\textit{Large-Scale Datasets:} For the large-scale reconstructions, we use
MatrixCity~\cite{li2023matrixcity}, Rubble4k~\cite{meganerf} and
Uni10k~\cite{windisch2026ALoDOfGaussians}. For MatrixCity and Uni10k, we use the
official test/train split. For Rubble4k, we use every 8th image as a test image.

\textit{Metrics:} We report PSNR, SSIM and LPIPS on the test views. For the
standard datasets, each number is averaged over $5$ seeds, and the dataset
averages weight every scene equally. Peak memory is measured with PyTorch's
memory profiler for the code of Faster-GS~\cite{hahlbohm2026fastergs}, and
training time is the total wall clock time of the $30$k iterations, excluding
data loading and evaluation. Following Faster-GS~\cite{hahlbohm2026fastergs} and
CLM~\cite{zhao2025clm}, we compute LPIPS with the VGG backbone, and normalize
the images to $[-1,1]$ before computing it.

\textit{Hardware:} We use a compute cluster for our experiments which injects
some additive noise in our time calculations due to potential contention with
other jobs. Our experiments use the following NVIDIA GPUs: Titan Xp, RTX 4090,
L40S and RTX PRO 6000 Blackwell. Each experiment states the GPU it uses.

\section{Densification Schemes}
\label{sec:supp_densification}
We use a variety of densification schemes to test our scenes due to two reasons:
(\rom{1}) schemes catered for small scale scenes fail to work at a larger scale,
(\rom{2}) we want to showcase our method is independent of the densification
scheme. All comparisons against baselines always use identical initializations
and densification schemes. The schemes we use are: adaptive density control as
described in the original 3DGS paper~\cite{kerbl3Dgaussians}, a budget-capped
version of it for the scaling run (described later), 3DGS-MCMC~\cite{3dgsmcmc},
and a modified densification scheme inspired from ``A LoD of
Gaussians''~\cite{windisch2026ALoDOfGaussians} (described later). Each
experiment states which densification scheme it uses.

\paragraph{Small Scale} For all the standard datasets, we first use adaptive
density control over $5$ seeds to find the average count of primitives that the
scenes converge to. We then use 3DGS-MCMC with this primitive count as the final
budget for comparing the optimizers and performing the ablation.

\paragraph{Scaling-Up Experiment} For the budget-capped adaptive density
control, the allowed Gaussian count rises linearly from the SfM count (iteration
$600$) to the target $N$ (iteration $12$k), and each densification step can only
add up to that cap. With the gradient threshold set to $0$, the Gaussians with
the largest mean view-space gradients are cloned or split first. The schedule
and the clone/split/prune rules are unchanged, so the budget, and not the
threshold, sets the final $N$. After iteration $12$k (the last opacity reset),
densification only replaces the pruned Gaussians. The purpose of this is to push
primitive density in the scene to test the training limits of the system rather
than place the primitives optimally.

\paragraph{Modified Densification at Large Scale} For some large scale
experiments, we use a modified densification scheme heavily inspired from ``A
LoD of Gaussians''~\cite{windisch2026ALoDOfGaussians} since it outperformed
naive adaptive density control in quality in our experiments, and we wanted to
test whether our optimizer performs on-par with Adam at high quality
reconstructions.
The densification happens every $1$k iterations from iteration $600$, with an
opacity reset to $\min(o, 0.01)$ happening every $3$k iterations, where $o$ is
the current opacity. Each Gaussian tracks a criterion---the maximum of the norm
of its view-space positional gradient over the iterations since the last
densification step, which is reset after every densification step. In each
densification step, Gaussians with opacity less than $0.005$ or a degenerate
rotation (norm less than $10^{-8}$) are declared dead, and every live Gaussian
whose criterion exceeds the gradient threshold for splitting ($0.0003$) is split
into two. The new Gaussians get the source Gaussian's opacity, rotation and
color but have a scale $1.6\times$ smaller.  Their position is offset from the
parent's mean by drawing randomly from $\mathcal{N}(0, R\,\text{diag}(s^2)R^T)$
and they start from a zeroed-out optimizer state.  We relocate the dead
Gaussians onto a live target sampled with replacement with probability
proportional to the opacity multiplied by $(1 + \text{criterion})$.  We adjust
the opacity and scale of the target and its copies with the 3DGS-MCMC relocation
formula. The number of Gaussians is limited by one hard global cap which is
incremented in steps of $5$ million until the GPU goes out of memory.

\section{Optimizer Implementation Details}
\label{sec:supp_implementation}
We use Adam for optimizing all of the parameters except the color parameters
(i.e., the SH coefficients), for which we use our proposed optimizer. Following
3DGS~\cite{kerbl3Dgaussians} and Faster-GS~\cite{hahlbohm2026fastergs}, the SH
coefficients are split into two tensors: one for the base diffuse band (DC) and
one for the specular bands, with learning rates of $2.5\times10^{-3}$ and
$1.25\times10^{-4}$ ($= 2.5\times10^{-3}/20$) respectively, the same as for
Adam. Both tensors share the single pooled second moment $\bar\nu$ of a
Gaussian, and only the step size differs between them. For the SH coefficients,
we do not allocate tensors for the first moment or for the gradient, and
allocate only an $N$ sized tensor for $\bar\nu$, where $N$ is the number of
Gaussians. We use the same $\beta_2$ and $\epsilon$ as Adam.  Like Adam, we
bias-correct $\bar\nu$ with $1-\beta_2^t$, using a single global step counter
$t$ that is shared by all the Gaussians. The $3$ DC primals are stored as fp32
and the remaining $45$ as fp16.

The fused kernel of the main paper runs right after the
\texttt{renderBackward} kernel, which provides the color gradient
$\frac{\partial L}{\partial c}$ of each Gaussian. A single kernel handles all the Gaussians. For
the visible ones, i.e., the Gaussians that touch at least one tile, it updates
$\bar\nu$ as in the main paper, evaluates the SH basis at the
viewing direction to compute the gradient of the SH coefficients
and updates the primals. For the invisible
ones, it only decays $\bar\nu$ by $\beta_2$.

The SH degree rises to $1$, $2$ and $3$ at iterations $1$k, $2$k and $3$k, i.e.,
the number of active bases $n$ goes from $1$ to $4$, $9$ and $16$. We apply
Uns\"old's theorem over the active bands only, so the sum of the squared SH
basis functions in the $\bar\nu$ update of the main paper is $n/4\pi$ instead of
$4/\pi$. While $n = 1$ (no specular coefficient active), $\bar\nu$ is still
updated every step, but no specular coefficient is written. Thus when band $1$
switches on, $\bar\nu$ is already warm, whereas Adam's second moment for these
coefficients would still be at zero. For $n > 1$, only the active bands are read
and written, and the inactive ones stay exactly zero.

\paragraph{Stochastic Rounding.} We find that large scale scenes that need to be trained for a lot of
iterations degrade in quality when training with the specular SH coefficients represented in
fp16. We find that using stochastic rounding while writing back the values helps
in recovering the quality back. To round a value stochastically, we add a random
$13$-bit number to the $13$ low bits of its fp32 mantissa, which are the bits
dropped when converting to fp16, and then truncate to fp16.  When fp16 values
are subnormal and evenly spaced, we instead compute the rounding probability
from the two fp16 neighbours explicitly and compare it against a uniform random
number. Algorithm~\ref{alg:fused_sh} summarizes our fused SH kernel.

\begin{algorithm}[t]
\caption{Fused SH kernel at step $t$, run right after
\texttt{renderBackward}. $K_{k,j} \in \mathbb{R}^3$ is the $j$-th SH coefficient
of Gaussian $k$, $o$ the camera center, $n$ the number of active bases, and
$\alpha_0$, $\alpha$ the learning rates of the DC and the specular
coefficients. $K_{k,0}$ is stored in fp32 and $K_{k,j>0}$ in fp16.
$\mathrm{SR}(\cdot)$ and $\mathrm{RN}(\cdot)$ round to fp16 stochastically and to
nearest, and the flag $\mathit{sr}$ turns stochastic rounding on.}
\label{alg:fused_sh}
\begin{algorithmic}[1]
\ForAll{Gaussians $k$ \textbf{in parallel}}
  \If{$k$ touches no tile}
    \State $\bar\nu_k \gets \beta_2\,\bar\nu_k$ \Comment{invisible: only decay}
    \State \textbf{return}
  \EndIf
  \State $g \gets \frac{\partial L}{\partial c_k}$ \Comment{from \texttt{renderBackward}}
  \State $\bar\nu_k \gets \beta_2\,\bar\nu_k + (1-\beta_2)\,\frac{1}{12\pi}\|g\|^2$
  \State $s \gets 1 / \big(\sqrt{\bar\nu_k / (1-\beta_2^t)} + \epsilon\big)$
  \State $K_{k,0} \gets K_{k,0} - \alpha_0\, s\, Y_0\, g$
  \If{$n > 1$}
    \State $d \gets (\mu_k - o) / \|\mu_k - o\|$
    \State $\frac{\partial L}{\partial \mu_k} \mathrel{+}= \big(\frac{\partial c_k}{\partial \mu_k}\big)^{\!\top} g$ \Comment{with $K_k$ before the step}
    \For{$j = 1, \dots, n-1$}
      \State $\tilde K \gets K_{k,j} - \alpha\, s\, Y_j(d)\, g$ \Comment{in fp32}
      \If{$\mathit{sr}$}
        \State $K_{k,j} \gets \mathrm{SR}(\tilde K)$
      \Else
        \State $K_{k,j} \gets \mathrm{RN}(\tilde K)$
      \EndIf
    \EndFor
  \EndIf
\EndFor
\end{algorithmic}
\end{algorithm}

\section{Teaser Experiment}
\label{sec:supp_teaser}
The teaser of the main paper compares the optimizers with equal amount of memory
available for training. It compares the \textsc{[Adam + bwd.\ fusion]} and
\textsc{[Ours + bwd.\ fusion]} for Spherical Harmonics on the
MatrixCity~\cite{li2023matrixcity} (small city, aerial views, all $10$ blocks)
on a single 12\,GB NVIDIA Titan Xp. The scene is initialized with identical
random subset of $2$M points from the ground truth point cloud that MatrixCity
provides. We train for $300$k iterations at batch size $1$ with the standard
3DGS training loss. For this run of MatrixCity, we used the modified
densification schedule (\S\ref{sec:supp_densification}) and also enable
stochastic rounding when writing back our $16$-bit SH primals. The final
primitive count was decided by using the per-Gaussian memory usage to predict
how many Gaussians the optimizer can fit while coming close to the memory
ceiling of the GPU.  Figure~\ref{fig:equal_mem_renders} shows more test views of
this run. Over all $741$ test views, ours has a higher PSNR on $93\%$ and a
lower LPIPS on $99.9\%$ of them. Since ours fits $2.5\times$ more Gaussians in
the same memory, it recovers fine details, e.g., building facades, that the $10$M
Gaussians of Adam are too few to represent. The gap comes from the primitive
count and not from the optimizer itself.

\section{Training Loss Trajectory}

\begin{figure*}
  \centering
  \includegraphics{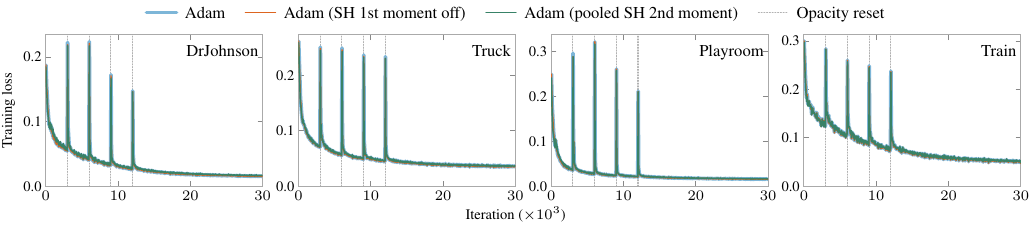}
  \caption{Training loss trajectory for different scenes of standard Adam, Adam using $\beta_1=0$ for all Spherical Harmonic (SH) coefficients, Adam using averaged per-Gaussian second memory for SH coefficients. We retain the standard Adam for all other parameter groups.}
  \label{fig:train_loss}
\end{figure*}

Figure~\ref{fig:train_loss} provides additional examples of the training loss of Adam, Adam with first moment dropped for SH coefficients, and Adam with second moment pooled/averaged for SH coefficients. For each experiment, we averaged across two seeds with consistent (but random) camera selection across the three optimizers. For these experiments, we used Faster-GS's adaptive density control (ADC). To reduce step-to-step noise and highlight the overall trend, we average the training loss over consecutive, non-overlapping blocks of 50 steps for each seed separately. We run these experiments on a NVIDIA GeForce RTX 4090 with 24 GB VRAM.

\section{Optimizer Comparison}

\paragraph{Learning Rate Search.} Since each baseline replaces Adam only for the
specular SH coefficients (see the main paper), we search the learning rate of only
these coefficients, and keep the default learning rates for all the other
parameters. For each baseline, we try $\{0.1, 0.3, 1, 3, 10\}\times$ its default
learning rate on all $11$ scenes, and keep the best one. We search Selective
Adam and our optimizer in the same way. RMSProp, Adam-mini, Selective Adam and
ours are best at their default, Adafactor at $10\times$, and Adafactor (HF) and
$8$-bit Adam at $3\times$. Adafactor is at the edge of the grid, but its PSNR
barely changes between $3\times$ and $10\times$. For SGD, we try every power of
$10$ from $10^0$ to $10^5$ and then $2\times10^2$ and $5\times10^2$;
$2\times10^2$ is the best, and $10^4$ and above diverge. The main paper reports
all of these at their best learning rate, which for our optimizer is the
default learning rate of Adam. ReducedSH is a variant of Adam and uses its
learning rates.

Tables~\ref{tab:supp-optimizer-per-scene} and~\ref{tab:supp-optimizer-per-scene-perf}
break down the dataset averages of the optimizer comparison in the main paper
into individual scenes for all optimizers, for image quality, and for memory and
training time. Our optimizer tracks
Adam on every scene, not just on average. Figure~\ref{fig:renders} shows a test
view of each of five scenes rendered by models trained with Adam and Ours, for
seed $0$. The renders of the two optimizers are visually indistinguishable.

\section{Scaling Up Behaviour}
\label{sec:supp_scaling}
To compare the optimizers at identical Gaussian counts, the resolution sweep of
the main paper uses the budget-capped ADC of \S\ref{sec:supp_densification}.
The final counts are within $0.1\%$ of $N$, and all runs use one L40S. We do not
report the quality, since it saturates and the hyperparameters are not tuned
for large Gaussian counts.
Also, \emph{bicycle} has only about $200$ training images while large scenes
have thousands, so the trend there may differ with a larger set of images
depending on whether the images are stored on host or device.

In the backward pass, Fused Adam from Faster-GS~\cite{hahlbohm2026fastergs} only
writes the gradients of the projected 2D Gaussians, and never stores the
gradients of the 3D primitives. So at scale its memory depends primarily on the
primitive count and the curves of all three resolutions meet in the resolution
sweep of the main paper. We do not store the Spherical Harmonics gradients
either, and shrink their optimizer states, i.e., the static memory, which lowers
the slope further.

\section{CLM Integration}
\label{sec:supp_clm}
The CLM experiments of the main paper train on MatrixCity (big city, aerial
views) for $500$k images at a batch size of $64$, i.e., $7{,}813$ iterations.
Following CLM~\cite{zhao2025clm}, the learning rates and the $\beta$s of Adam
are scaled with the batch size as in Grendel~\cite{grendelgs}, i.e., the
learning rates by its square root and, e.g., $\beta_1 = 0.9^{64}$. Since CLM uses
Selective Adam~\cite{10.1145/3680528.3687694}, we also use a selective variant
of our optimizer, which does not update the pooled second moment $\bar\nu$ of
the invisible Gaussians, unlike \S\ref{sec:supp_implementation} where it is
decayed by $\beta_2$. Densification is turned off, and the primitive count is
set by the initialization: the $102$M run starts from the official point cloud
of MatrixCity, and the $25$M and $51$M runs from a random $25\%$ and $50\%$
subsample of it. Each configuration is a single run on one L40S, with $16$ CPU
cores on the same socket as the GPU. We do not use stochastic rounding for the
fp16 SH primals in these runs because we do not see a quality degradation with
fp16 with CLM. We compute PSNR and SSIM in floating point over all $9{,}066$
test views at the native resolution with a black background, rather than on
8-bit images as in Faster-GS~\cite{hahlbohm2026fastergs}, and LPIPS uses the VGG
backbone. The training time is the total wall clock time of the run. The PCIe
traffic and the peak GPU memory are measured with NVML.
Figure~\ref{fig:clm_renders} shows a test view rendered at each of the three
budgets, a different view for each. The renders of Selective Adam and ours are
visually indistinguishable.

\section{Grendel Integration}
\label{sec:supp_grendel}

Grendel~\cite{grendelgs} is built on the original 3DGS
codebase~\cite{kerbl3Dgaussians}, and its rasterizer lacks the speed-ups
that recent work brought to 3DGS training. Measuring our optimizer against it
would compare against a baseline that is slower and uses more memory than
necessary. We therefore first port the speed-ups of
Faster-GS~\cite{hahlbohm2026fastergs} into Grendel, and report all Grendel
results, for both Adam and ours, on this updated codebase.

We only port features that do not change the training math: the rendered image
is unchanged, and the gradients differ only in the order of floating point
summation, which Grendel's atomics already make nondeterministic. We leave out
the changes of Faster-GS that alter training, e.g., culling degenerate
Gaussians, removing the $0.99$ clamp on $\alpha$, or skipping updates of
invisible Gaussians, as well as its separate depth and tile sorting and its
z-order sorting of Gaussians. The ported features are:
\begin{itemize}
  \item \textit{Exact tile culling.} Instead of every tile in a square of
  radius $3\sigma$, a Gaussian is assigned only to the tiles where it can reach
  $\alpha \geq 1/255$~\cite{hahlbohm2026fastergs, radl2024stopthepop}. This
  shortens the per-tile lists of Gaussians, and since Grendel sends a Gaussian
  only to the GPUs whose tiles it touches, also reduces the communication.
  \item \textit{Per-Gaussian backward pass.} We replace the per-pixel backward
  pass with the per-Gaussian one of Taming-3DGS~\cite{10.1145/3680528.3687694}
  and Faster-GS. To compute the gradients, they obtain the color behind a
  Gaussian by subtracting the color accumulated in front of it from the final
  pixel color, which loses precision deep in a pixel's list. We instead sum the
  blended colors of the Gaussians behind it. The forward pass is unchanged.
  \item \textit{Compact preprocessing.} For every view in a batch, Grendel
  keeps the projected attributes (2D mean, conic, color) and the gradients of
  all the Gaussians on a GPU, although only a small fraction of them is visible
  in a view. We keep only the visible Gaussians of each view, and accumulate
  their gradients into one buffer per batch.
  \item \textit{Separate SH buffers.} As in
  Taming-3DGS~\cite{10.1145/3680528.3687694} and
  Faster-GS~\cite{hahlbohm2026fastergs}, the DC and the specular SH
  coefficients go to the rasterizer as two tensors instead of being concatenated
  at every batch.
  \item \textit{Fused kernels.} PyTorch's fused Adam, as in Faster-GS, and the
  fused SSIM kernel of Taming-3DGS~\cite{10.1145/3680528.3687694} with the
  same window and constants.
\end{itemize}
The update makes training with Adam $4.6\times$ faster on MatrixCity on $4$
GPUs.

All Grendel runs use $4$ RTX PRO 6000 Blackwell GPUs. Grendel counts
iterations in images, not in optimizer steps, and we do the same here. For
our optimizer, the SH step covers both the DC and the specular SH coefficients, and
the positions, scales, rotations and opacities stay on Adam, as in
\S\ref{sec:supp_implementation}. Our SH step runs inside the backward pass by
construction, but unlike Faster-GS, we do not fuse the Adam step of the other
parameters into it. The baseline uses Adam for all parameters.
PSNR and SSIM are computed on 8-bit images over all test views with the metric
code of Grendel. Unlike \S\ref{sec:supp_experimental_details}, we follow the
codebase of Grendel~\cite{grendelgs} and pass the images to LPIPS in $[0,1]$
instead of $[-1,1]$. The memory reported in the main paper is the peak of the
training phase on the worst GPU, and the time excludes the test passes.

\paragraph{Rubble4k and Uni10k.} Both train with a batch size of $1$ for $200$k
iterations, with the modified densification of \S\ref{sec:supp_densification},
but only until iteration $50$k, capped at $28.0$M Gaussians for Rubble4k and
$28.6$M for Uni10k. For Rubble4k, we additionally prune every $1$k iterations
the Gaussians whose largest scale exceeds $0.1\times$ the camera extent, as in
Grendel. For Uni10k, the position learning rate decays over $30$k iterations
and the opacity learning rate is $0.05$.

\paragraph{MatrixCity:} We train on the small city (aerial views, all blocks) with
Grendel's recipe: a batch size of $16$ for $300$k iterations, i.e., $18.75$k
optimizer steps, with the learning rates scaled by the square root of the batch
size (\S\ref{sec:supp_clm}), starting from the ground-truth point cloud of
MatrixCity downsampled to $15.5$M points. We use the adaptive density control
of Grendel from iteration $20$k, every $500$ iterations, with a gradient
threshold of $2\times10^{-4}$ and an opacity reset every $45$k iterations. We
stop densifying at iteration $75{,}489$, where Adam reaches the $36$\,GB memory
limit, the default guard of Grendel, so that both optimizers end at $36.6$M Gaussians. We score all $741$ test
views with a separate render pass, since the evaluation of Grendel drops the
last few views of a batch.

\paragraph{Batch Size Larger Than One.} Our update relies on the SH gradient
being rank-1, which is true for a single view but not for a sum over several
views. Distributed trainers such as Grendel, however, average gradients over a
batch of views before each optimizer step. We sidestep this by never summing the
SH gradient. Instead, we take one SH update per view, inside that view's
backward pass, so each step sees an exact rank-1 gradient. A batch of $B$ views
thus becomes $B$ small sequential steps rather than one large averaged step,
which is the behaviour that batch-size scaling rules try to imitate in the first
place. Only the SH coefficients are stepped this way; all other parameters keep
the trainer's usual batched update.

\providecolor{linkblue}{rgb}{0.21,0.49,0.74}
\colorlet{stagefwd}{linkblue!35}
\colorlet{stagebo}{orange!45}
\pgfplotstableread{
x   fwd   bo
0   2.07  0.64
1   1.94  0.76
2   3.20  1.56
3   3.31  1.45
5   3.54  8.43
6   3.38  11.40
7   4.01  7.36
8   3.27  4.79
10  4.45  16.17
11  3.96  14.91
12  4.12  8.36
13  3.28  4.82
16  2.11  0.82
17  2.01  0.92
18  3.29  1.73
19  3.42  1.60
21  3.19  3.21
22  3.49  4.45
23  4.12  4.01
24  3.33  2.58
26  4.51  6.05
27  4.10  5.76
28  4.37  4.49
29  3.41  2.58
}\rttimes
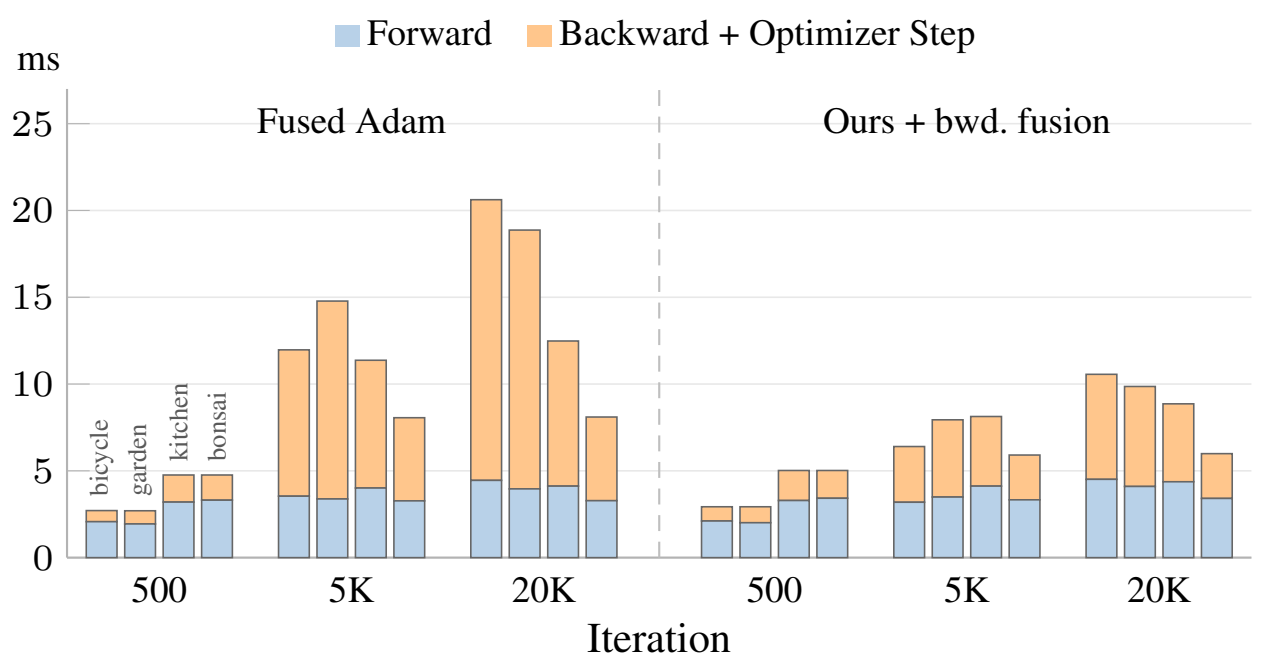
\begin{figure}
\centering
\begin{tikzpicture}
\begin{axis}[
  ybar stacked, bar width=0.8,
  scale only axis=true, width=0.94\columnwidth, height=3.1cm,
  xmin=-0.9, xmax=29.9, ymin=0, ymax=27,
  axis lines*=left, axis line style={gray!60, line width=0.4pt},
  ymajorgrids=true, grid style={gray!18, line width=0.3pt},
  ytick={0,5,10,15,20,25}, tick style={gray!60}, yticklabel style={xshift=1pt},
  xtick={1.5,6.5,11.5,17.5,22.5,27.5}, major x tick style={draw=none},
  xticklabels={500,5K,20K,500,5K,20K},
  tick label style={font=\scriptsize}, label style={font=\footnotesize},
  xlabel={Iteration}, xlabel style={yshift=3pt},
  ylabel={ms}, ylabel style={rotate=-90, font=\scriptsize,
    at={(axis description cs:0,1)}, anchor=south east, inner xsep=1pt},
  every axis plot/.append style={draw=black!60, line width=0.3pt},
  legend style={font=\scriptsize, draw=none, fill=none,
    at={(0.5,1.03)}, anchor=south, legend columns=-1,
    /tikz/every even column/.append style={column sep=5pt},
    nodes={inner xsep=1.5pt}},
  legend image code/.code={\fill[#1, draw=none] (0cm,-0.07cm) rectangle (0.16cm,0.07cm);},
  legend cell align=left,
  clip=false,
]
\addplot[fill=stagefwd] table[x=x, y=fwd] {\rttimes};
\addlegendentry{Forward}
\addplot[fill=stagebo] table[x=x, y=bo] {\rttimes};
\addlegendentry{Backward + Optimizer Step}

\draw[gray!50, densely dashed, line width=0.4pt]
  (axis cs:14.5,0) -- (axis cs:14.5,27);
\node[font=\scriptsize, anchor=north] at (axis cs:6.5,27) {Fused Adam};
\node[font=\scriptsize, anchor=north] at (axis cs:22.5,27) {Ours + bwd.\ fusion};

\tikzset{scene/.style={font=\tiny, rotate=90, anchor=west, inner sep=1pt,
  text=black!65}}
\node[scene] at (axis cs:0,3.2) {bicycle};
\node[scene] at (axis cs:1,3.2) {garden};
\node[scene] at (axis cs:2,5.2) {kitchen};
\node[scene] at (axis cs:3,5.2) {bonsai};

\end{axis}
\end{tikzpicture}
\caption{Breakdown of the time of a training iteration for Fused Adam and
Ours + bwd.\ fusion on four scenes at three points of training, measured on one
NVIDIA L40S. Backward fusion is enabled for both, so the
backward pass and the optimizer step are shown together.}
\label{fig:runtime_breakdown}
\end{figure}

\section{Ablation Study}
The ablation uses the same setup as the optimizer comparison: the $11$
standard scenes, $5$ seeds, and 3DGS-MCMC with the budget from adaptive density
control (\S\ref{sec:supp_densification}). All ablation runs use one L40S.

Table~\ref{tab:supp-ablation-per-scene} breaks down the ablation of
the main paper into individual scenes.

To understand where the savings come from,
Figure~\ref{fig:runtime_breakdown} breaks down the time of a single training
iteration of the Fused Adam of Faster-GS~\cite{hahlbohm2026fastergs} and
Ours + bwd.\ fusion into the forward pass
and the backward pass with the optimizer step. Since backward fusion
(see the main paper) is enabled for both, we show the backward pass
and the optimizer step together. Unlike the other experiments, this one
uses the ADC densification of Faster-GS instead of MCMC, to match the runtime
breakdown of Faster-GS~\cite{hahlbohm2026fastergs}. The
forward pass takes about the same time for both, while, summed over the four
scenes, our fused SH kernel makes the backward pass and the optimizer step
together $2.3\times$ faster at iteration $20$k, leading to a $1.7\times$ faster
training iteration. Early in training, when there are only few Gaussians, ours
is slightly slower ($0.94\times$ at iteration $500$). The breakdown of the static memory is in
the main paper.

\clearpage
\providecommand{\tbd}{\textcolor{red}{TBD}}
\providecommand{\fusedrow}{\hspace{0.6em}\textcolor{gray}{$\hookrightarrow$}\,+ bwd.\ fusion}
\begin{table*}[t]
        \centering
        \footnotesize
        \setlength{\tabcolsep}{3pt}
        \renewcommand{\arraystretch}{1.0}
        \setlength{\aboverulesep}{0pt}
        \setlength{\belowrulesep}{0pt}
        \makebox[\textwidth][c]{\begin{tabular}{c@{\hskip 4pt}|l ccccccc cc cc c}
        \toprule
 & & \multicolumn{7}{c}{Mip-NeRF 360~\cite{barron2022mipnerf360}} & \multicolumn{2}{c}{Tanks \& Temples~\cite{Knapitsch2017}} & \multicolumn{2}{c}{Deep Blending~\cite{DeepBlending2018}} & \\
        \cmidrule(lr){3-9} \cmidrule(lr){10-11} \cmidrule(lr){12-13}
 & Method & bicycle & garden & stump & room & counter & kitchen & bonsai & \makebox[33pt]{train} & \makebox[33pt]{truck} & drjohnson & playroom & Avg \\
        \midrule
 & Budget (M)             & 4.81 & 4.14 & 4.20 & 1.30 & 1.06 & 1.59 & 1.07 & 1.06 & 2.02 & 3.22 & 1.90 & 2.40 \\
        \midrule
 & Adam                   & 25.63 & 27.64 & 27.13 & 32.29 & 29.39 & 31.80 & 32.54 & 22.70 & 26.33 & 29.49 & 30.09 & 28.64 \\
 & \fusedrow              & 25.64 & 27.65 & 27.18 & 32.30 & 29.41 & 31.73 & 32.54 & 22.71 & 26.32 & 29.41 & 30.06 & 28.63 \\
 & RMSProp      & 25.63 & 27.65 & 27.14 & 32.37 & 29.41 & 31.76 & 32.59 & 22.70 & 26.38 & 29.47 & 30.19 & 28.66 \\
 & Adam-mini    & 25.63 & 27.62 & 27.19 & 32.23 & 29.36 & 31.70 & 32.46 & 22.64 & 26.32 & 29.49 & 30.09 & 28.61 \\
 & Adafactor    & 25.68 & 27.61 & 27.15 & 31.88 & 29.28 & 31.54 & 32.10 & 22.63 & 26.27 & 29.37 & 29.91 & 28.49 \\
 & Adafactor (HF)& 25.61 & 27.68 & 27.04 & 32.23 & 29.61 & 31.78 & 32.64 & 22.82 & 26.42 & 29.42 & 29.97 & 28.66 \\
 & 8-bit Adam   & 25.59 & 27.67 & 26.71 & 32.42 & 29.61 & 31.91 & 32.69 & 22.52 & 26.20 & 29.58 & 30.26 & 28.65 \\
 & SGD          & 25.50 & 27.45 & 27.20 & 32.03 & 29.12 & 31.57 & 32.08 & 22.78 & 26.26 & 29.44 & 30.04 & 28.50 \\
 & Selective Adam & 25.28 & 27.58 & 26.33 & 32.23 & 29.23 & 31.83 & 32.44 & 22.46 & 26.12 & 29.36 & 30.36 & 28.47 \\
 & ReducedSH              & 25.62 & 27.58 & 27.21 & 32.33 & 29.32 & 31.65 & 32.54 & 22.43 & 26.31 & 29.63 & 30.17 & 28.62 \\
        \cmidrule{2-14}
 & Ours         & 25.63 & 27.61 & 27.21 & 32.23 & 29.34 & 31.74 & 32.46 & 22.67 & 26.37 & 29.49 & 30.31 & 28.64 \\
\multirow{-12}{*}{\rotatebox[origin=c]{90}{PSNR}} & \fusedrow              & 25.61 & 27.56 & 27.21 & 32.27 & 29.36 & 31.78 & 32.46 & 22.65 & 26.35 & 29.46 & 30.06 & 28.62 \\
        \midrule
 & Adam                   & 0.792 & 0.870 & 0.800 & 0.930 & 0.917 & 0.932 & 0.947 & 0.833 & 0.896 & 0.903 & 0.908 & 0.884 \\
 & \fusedrow              & 0.792 & 0.870 & 0.801 & 0.930 & 0.917 & 0.932 & 0.947 & 0.833 & 0.896 & 0.902 & 0.908 & 0.884 \\
 & RMSProp      & 0.792 & 0.870 & 0.800 & 0.930 & 0.917 & 0.932 & 0.947 & 0.834 & 0.896 & 0.902 & 0.909 & 0.884 \\
 & Adam-mini    & 0.791 & 0.869 & 0.802 & 0.930 & 0.916 & 0.932 & 0.946 & 0.832 & 0.896 & 0.903 & 0.908 & 0.884 \\
 & Adafactor    & 0.794 & 0.868 & 0.802 & 0.928 & 0.915 & 0.930 & 0.944 & 0.831 & 0.895 & 0.901 & 0.909 & 0.883 \\
 & Adafactor (HF)& 0.790 & 0.870 & 0.797 & 0.930 & 0.918 & 0.932 & 0.947 & 0.834 & 0.896 & 0.902 & 0.908 & 0.884 \\
 & 8-bit Adam   & 0.785 & 0.868 & 0.783 & 0.930 & 0.917 & 0.932 & 0.947 & 0.826 & 0.892 & 0.904 & 0.909 & 0.881 \\
 & SGD          & 0.789 & 0.866 & 0.803 & 0.929 & 0.913 & 0.929 & 0.944 & 0.833 & 0.895 & 0.902 & 0.910 & 0.883 \\
 & Selective Adam & 0.772 & 0.867 & 0.770 & 0.928 & 0.911 & 0.931 & 0.944 & 0.821 & 0.891 & 0.900 & 0.909 & 0.877 \\
 & ReducedSH              & 0.793 & 0.869 & 0.803 & 0.930 & 0.916 & 0.931 & 0.947 & 0.831 & 0.896 & 0.903 & 0.908 & 0.884 \\
        \cmidrule{2-14}
 & Ours         & 0.792 & 0.869 & 0.801 & 0.930 & 0.916 & 0.931 & 0.946 & 0.833 & 0.896 & 0.903 & 0.909 & 0.884 \\
\multirow{-12}{*}{\rotatebox[origin=c]{90}{SSIM}} & \fusedrow              & 0.792 & 0.869 & 0.801 & 0.930 & 0.916 & 0.932 & 0.946 & 0.833 & 0.896 & 0.902 & 0.908 & 0.884 \\
        \midrule
 & Adam                   & 0.194 & 0.118 & 0.211 & 0.255 & 0.237 & 0.149 & 0.235 & 0.221 & 0.130 & 0.303 & 0.297 & 0.214 \\
 & \fusedrow              & 0.194 & 0.118 & 0.211 & 0.255 & 0.237 & 0.149 & 0.236 & 0.221 & 0.131 & 0.304 & 0.297 & 0.214 \\
 & RMSProp      & 0.195 & 0.118 & 0.212 & 0.255 & 0.236 & 0.149 & 0.235 & 0.220 & 0.131 & 0.303 & 0.298 & 0.214 \\
 & Adam-mini    & 0.195 & 0.119 & 0.210 & 0.256 & 0.237 & 0.149 & 0.236 & 0.222 & 0.131 & 0.303 & 0.298 & 0.214 \\
 & Adafactor    & 0.191 & 0.120 & 0.210 & 0.257 & 0.239 & 0.150 & 0.237 & 0.226 & 0.131 & 0.306 & 0.299 & 0.215 \\
 & Adafactor (HF)& 0.195 & 0.118 & 0.215 & 0.255 & 0.235 & 0.148 & 0.235 & 0.219 & 0.130 & 0.304 & 0.299 & 0.214 \\
 & 8-bit Adam   & 0.208 & 0.122 & 0.231 & 0.260 & 0.238 & 0.150 & 0.241 & 0.230 & 0.142 & 0.305 & 0.311 & 0.222 \\
 & SGD          & 0.198 & 0.123 & 0.210 & 0.257 & 0.241 & 0.151 & 0.238 & 0.222 & 0.131 & 0.304 & 0.298 & 0.216 \\
 & Selective Adam & 0.218 & 0.123 & 0.242 & 0.264 & 0.249 & 0.151 & 0.245 & 0.235 & 0.147 & 0.308 & 0.308 & 0.226 \\
 & ReducedSH              & 0.192 & 0.119 & 0.208 & 0.255 & 0.238 & 0.149 & 0.235 & 0.223 & 0.132 & 0.301 & 0.297 & 0.214 \\
        \cmidrule{2-14}
 & Ours         & 0.197 & 0.119 & 0.212 & 0.255 & 0.237 & 0.149 & 0.236 & 0.221 & 0.131 & 0.305 & 0.300 & 0.215 \\
\multirow{-12}{*}{\rotatebox[origin=c]{90}{LPIPS}} & \fusedrow              & 0.197 & 0.119 & 0.212 & 0.255 & 0.237 & 0.149 & 0.236 & 0.220 & 0.131 & 0.305 & 0.301 & 0.215 \\
        \bottomrule
        \end{tabular}}%
        \caption{Per-scene image quality for the optimizer comparison of
        the main paper. Each cell is averaged over 5
        seeds. \textit{Avg} weights the $11$ scenes equally. \textit{Budget}:
        primitive budget per scene (millions), set to the primitive count
        that the original 3DGS ADC converges to.
        \protect$\hookrightarrow$\,\textit{+ bwd.\ fusion}: the update of all
        parameters of the visible Gaussians is fused into the backward pass, as
        in the Fused Adam of Faster-GS~\cite{hahlbohm2026fastergs}.
        Each optimizer either uses the learning rate proposed for 3DGS, or one
        tuned with a grid search for the highest average quality over the $11$
        scenes. For SGD the search is
        necessary, otherwise it diverges, since unlike Adam, its update is not
        normalized by the gradient magnitude. Higher PSNR and SSIM, and lower LPIPS are better.}
        \label{tab:supp-optimizer-per-scene}
        \end{table*}

\begin{table*}[t]
        \centering
        \footnotesize
        \setlength{\tabcolsep}{3pt}
        \renewcommand{\arraystretch}{1.0}
        \setlength{\aboverulesep}{0pt}
        \setlength{\belowrulesep}{0pt}
        \makebox[\textwidth][c]{\begin{tabular}{c@{\hskip 4pt}|l ccccccc cc cc c}
        \toprule
 & & \multicolumn{7}{c}{Mip-NeRF 360~\cite{barron2022mipnerf360}} & \multicolumn{2}{c}{Tanks \& Temples~\cite{Knapitsch2017}} & \multicolumn{2}{c}{Deep Blending~\cite{DeepBlending2018}} & \\
        \cmidrule(lr){3-9} \cmidrule(lr){10-11} \cmidrule(lr){12-13}
 & Method & bicycle & garden & stump & room & counter & kitchen & bonsai & \makebox[33pt]{train} & \makebox[33pt]{truck} & drjohnson & playroom & Avg \\
        \midrule
 & Budget (M)             & 4.81 & 4.14 & 4.20 & 1.30 & 1.06 & 1.59 & 1.07 & 1.06 & 2.02 & 3.22 & 1.90 & 2.40 \\
        \midrule
 & Adam                   & 5.83 & 4.78 & 5.09 & 2.28 & 1.95 & 2.38 & 1.58 & 1.47 & 2.57 & 3.98 & 2.31 & 3.11 \\
 & \fusedrow              & 4.86 & 4.18 & 4.24 & 2.03 & 1.74 & 2.00 & 1.37 & 1.20 & 2.09 & 3.28 & 1.92 & 2.63 \\
 & RMSProp      & 5.02 & 4.08 & 4.38 & 2.07 & 1.77 & 2.12 & 1.39 & 1.29 & 2.23 & 3.43 & 2.00 & 2.71 \\
 & Adam-mini    & 5.06 & 4.10 & 4.40 & 1.98 & 1.75 & 2.15 & 1.41 & 1.30 & 2.24 & 3.45 & 2.00 & 2.71 \\
 & Adafactor    & 4.25 & 3.39 & 3.71 & 1.86 & 1.62 & 1.90 & 1.27 & 1.12 & 1.91 & 2.94 & 1.70 & 2.33 \\
 & Adafactor (HF)& 4.24 & 3.40 & 3.73 & 1.81 & 1.62 & 1.87 & 1.24 & 1.12 & 1.90 & 2.93 & 1.69 & 2.32 \\
 & 8-bit Adam   & 4.41 & 3.60 & 3.91 & 1.90 & 1.55 & 1.96 & 1.22 & 1.09 & 2.01 & 3.21 & 1.88 & 2.43 \\
 & SGD          & 4.18 & 3.38 & 3.67 & 1.70 & 1.46 & 1.88 & 1.26 & 1.12 & 1.89 & 2.91 & 1.65 & 2.28 \\
 & Selective Adam & 5.48 & 4.67 & 4.99 & 1.96 & 1.74 & 2.45 & 1.48 & 1.40 & 2.41 & 4.12 & 2.39 & 3.01 \\
 & ReducedSH              & 6.62 & 5.46 & 5.80 & 2.45 & 2.14 & 2.67 & 1.77 & 1.65 & 2.90 & 4.54 & 2.63 & 3.51 \\
        \cmidrule{2-14}
 & Ours         & 2.92 & 2.25 & 2.54 & 1.46 & 1.24 & 1.38 & 0.92 & 0.81 & 1.34 & 2.02 & 1.17 & 1.64 \\
\multirow{-12}{*}{\rotatebox[origin=c]{90}{VRAM (GB)}} & \fusedrow              & 2.68 & 2.04 & 2.32 & 1.44 & 1.25 & 1.34 & 0.90 & 0.76 & 1.24 & 1.86 & 1.08 & 1.54 \\
        \midrule
 & Adam                   & 647 & 634 & 585 & 364 & 375 & 460 & 337 & 230 & 377 & 440 & 289 & 431 \\
 & \fusedrow              & 612 & 604 & 556 & 350 & 366 & 446 & 327 & 230 & 370 & 379 & 257 & 409 \\
 & RMSProp      & 591 & 578 & 529 & 338 & 354 & 432 & 313 & 212 & 344 & 384 & 262 & 394 \\
 & Adam-mini    & 591 & 583 & 529 & 335 & 357 & 432 & 312 & 211 & 346 & 383 & 261 & 395 \\
 & Adafactor    & 760 & 735 & 669 & 400 & 407 & 507 & 364 & 263 & 428 & 509 & 336 & 489 \\
 & Adafactor (HF)& 643 & 629 & 571 & 363 & 369 & 456 & 333 & 228 & 370 & 423 & 285 & 425 \\
 & 8-bit Adam   & 567 & 564 & 508 & 328 & 354 & 451 & 301 & 199 & 335 & 343 & 246 & 381 \\
 & SGD          & 528 & 524 & 471 & 314 & 331 & 408 & 291 & 197 & 316 & 334 & 230 & 359 \\
 & Selective Adam & 369 & 421 & 354 & 275 & 291 & 388 & 254 & 177 & 246 & 280 & 209 & 297 \\
 & ReducedSH              & 536 & 530 & 488 & 329 & 341 & 402 & 295 & 200 & 320 & 352 & 241 & 367 \\
        \cmidrule{2-14}
 & Ours         & 396 & 387 & 362 & 270 & 298 & 347 & 252 & 156 & 238 & 243 & 179 & 284 \\
\multirow{-12}{*}{\rotatebox[origin=c]{90}{Time (s)}} & \fusedrow              & 356 & 351 & 330 & 260 & 287 & 331 & 240 & 146 & 223 & 220 & 165 & 264 \\
        \bottomrule
        \end{tabular}}%
        \caption{Per-scene peak GPU memory and training time for the optimizer
        comparison of the main paper, for the same runs as
        Table~\ref{tab:supp-optimizer-per-scene}, on one L40S GPU.
        Each cell is averaged over 5 seeds. \textit{Avg} weights the $11$
        scenes equally. \textit{Budget}: primitive budget per scene
        (millions), set to the primitive count that the original 3DGS ADC
        converges to. \protect$\hookrightarrow$\,\textit{+ bwd.\ fusion}: the
        update of all parameters of the visible Gaussians is fused into the
        backward pass, as in the Fused Adam of
        Faster-GS~\cite{hahlbohm2026fastergs}.
        \textit{VRAM}: peak GPU memory (GB); \textit{Time}: total training time (s). Lower is better for both.}
        \label{tab:supp-optimizer-per-scene-perf}
        \end{table*}

\begin{table*}[t]
        \centering
        \small
        \setlength{\tabcolsep}{3pt}
        \makebox[\textwidth][c]{\begin{tabular}{l ccccccc cc cc c}
        \toprule
 & \multicolumn{7}{c}{Mip-NeRF 360~\cite{barron2022mipnerf360}} & \multicolumn{2}{c}{Tanks \& Temples~\cite{Knapitsch2017}} & \multicolumn{2}{c}{Deep Blending~\cite{DeepBlending2018}} & \\
        \cmidrule(lr){2-8} \cmidrule(lr){9-10} \cmidrule(lr){11-12}
Variant & bicycle & garden & stump & room & counter & kitchen & bonsai & \makebox[33pt]{train} & \makebox[33pt]{truck} & drjohnson & playroom & Avg \\
        \midrule
\multicolumn{13}{l}{\textit{PSNR$\uparrow$}} \\
Adam                   & 25.63 & 27.64 & 27.13 & 32.29 & 29.39 & 31.80 & 32.54 & 22.70 & 26.33 & 29.49 & 30.09 & 28.64 \\
+ drop $m$             & 25.64 & 27.63 & 27.20 & 32.31 & 29.44 & 31.81 & 32.60 & 22.76 & 26.32 & 29.49 & 30.28 & 28.68 \\
+ pool $\nu$           & 25.63 & 27.59 & 27.18 & 32.20 & 29.35 & 31.72 & 32.48 & 22.65 & 26.37 & 29.46 & 30.00 & 28.60 \\
+ fused SH kernel  & 25.64 & 27.58 & 27.23 & 32.21 & 29.35 & 31.73 & 32.47 & 22.60 & 26.37 & 29.53 & 30.06 & 28.61 \\
+ 16-bit primal        & 25.63 & 27.61 & 27.21 & 32.23 & 29.34 & 31.74 & 32.46 & 22.67 & 26.37 & 29.49 & 30.31 & 28.64 \\
+ bwd.\ fusion     & 25.61 & 27.56 & 27.21 & 32.27 & 29.36 & 31.78 & 32.46 & 22.65 & 26.35 & 29.46 & 30.06 & 28.62 \\
        \midrule
\multicolumn{13}{l}{\textit{SSIM$\uparrow$}} \\
Adam                   & 0.792 & 0.870 & 0.800 & 0.930 & 0.917 & 0.932 & 0.947 & 0.833 & 0.896 & 0.903 & 0.908 & 0.884 \\
+ drop $m$             & 0.792 & 0.870 & 0.801 & 0.930 & 0.917 & 0.932 & 0.947 & 0.834 & 0.896 & 0.902 & 0.909 & 0.885 \\
+ pool $\nu$           & 0.792 & 0.869 & 0.801 & 0.928 & 0.916 & 0.932 & 0.946 & 0.832 & 0.896 & 0.902 & 0.906 & 0.884 \\
+ fused SH kernel  & 0.791 & 0.869 & 0.802 & 0.930 & 0.916 & 0.932 & 0.946 & 0.833 & 0.896 & 0.902 & 0.906 & 0.884 \\
+ 16-bit primal        & 0.792 & 0.869 & 0.801 & 0.930 & 0.916 & 0.931 & 0.946 & 0.833 & 0.896 & 0.903 & 0.909 & 0.884 \\
+ bwd.\ fusion     & 0.792 & 0.869 & 0.801 & 0.930 & 0.916 & 0.932 & 0.946 & 0.833 & 0.896 & 0.902 & 0.908 & 0.884 \\
        \midrule
\multicolumn{13}{l}{\textit{LPIPS$\downarrow$}} \\
Adam                   & 0.194 & 0.118 & 0.211 & 0.255 & 0.237 & 0.149 & 0.235 & 0.221 & 0.130 & 0.303 & 0.297 & 0.214 \\
+ drop $m$             & 0.196 & 0.118 & 0.212 & 0.255 & 0.236 & 0.148 & 0.235 & 0.220 & 0.131 & 0.305 & 0.300 & 0.214 \\
+ pool $\nu$           & 0.197 & 0.119 & 0.212 & 0.255 & 0.237 & 0.149 & 0.236 & 0.222 & 0.131 & 0.305 & 0.301 & 0.215 \\
+ fused SH kernel  & 0.197 & 0.119 & 0.211 & 0.255 & 0.237 & 0.149 & 0.236 & 0.221 & 0.131 & 0.305 & 0.301 & 0.215 \\
+ 16-bit primal        & 0.197 & 0.119 & 0.212 & 0.255 & 0.237 & 0.149 & 0.236 & 0.221 & 0.131 & 0.305 & 0.300 & 0.215 \\
+ bwd.\ fusion     & 0.197 & 0.119 & 0.212 & 0.255 & 0.237 & 0.149 & 0.236 & 0.220 & 0.131 & 0.305 & 0.301 & 0.215 \\
        \midrule
\multicolumn{13}{l}{\textit{VRAM (GB)$\downarrow$}} \\
Adam                   & 5.83 & 4.78 & 5.09 & 2.28 & 1.95 & 2.38 & 1.58 & 1.47 & 2.57 & 3.98 & 2.31 & 3.11 \\
+ drop $m$             & 4.97 & 4.03 & 4.34 & 1.92 & 1.70 & 2.10 & 1.38 & 1.28 & 2.21 & 3.40 & 1.98 & 2.67 \\
+ pool $\nu$           & 4.13 & 3.30 & 3.60 & 1.80 & 1.53 & 1.82 & 1.19 & 1.09 & 1.86 & 2.85 & 1.65 & 2.26 \\
+ fused SH kernel  & 3.33 & 2.61 & 2.89 & 1.60 & 1.39 & 1.54 & 1.03 & 0.91 & 1.52 & 2.30 & 1.36 & 1.86 \\
+ 16-bit primal        & 2.92 & 2.25 & 2.54 & 1.46 & 1.24 & 1.38 & 0.92 & 0.81 & 1.34 & 2.02 & 1.17 & 1.64 \\
+ bwd.\ fusion     & 2.68 & 2.04 & 2.32 & 1.44 & 1.25 & 1.34 & 0.90 & 0.76 & 1.24 & 1.86 & 1.08 & 1.54 \\
        \midrule
\multicolumn{13}{l}{\textit{Time (s)$\downarrow$}} \\
Adam                   & 647 & 634 & 585 & 364 & 375 & 460 & 337 & 230 & 377 & 440 & 289 & 431 \\
+ drop $m$             & 595 & 579 & 534 & 340 & 359 & 436 & 309 & 209 & 346 & 383 & 264 & 396 \\
+ pool $\nu$           & 527 & 518 & 475 & 319 & 338 & 405 & 293 & 195 & 314 & 338 & 233 & 360 \\
+ fused SH kernel  & 404 & 398 & 376 & 280 & 306 & 356 & 258 & 159 & 247 & 248 & 181 & 292 \\
+ 16-bit primal        & 396 & 387 & 362 & 270 & 298 & 347 & 252 & 156 & 238 & 243 & 179 & 284 \\
+ bwd.\ fusion     & 356 & 351 & 330 & 260 & 287 & 331 & 240 & 146 & 223 & 220 & 165 & 264 \\
        \bottomrule
        \end{tabular}}%
        \caption{Per-scene results for the ablation of
        the main paper, with the second moment pooled over all $48$
        Spherical Harmonics coefficients. The last row adds backward fusion of the remaining geometry
        parameters. Each cell is averaged over 5 seeds. \textit{Avg}
        weights the $11$ scenes equally and corresponds to
        the main paper. \textit{VRAM}: peak GPU memory (GB);
        \textit{Time}: total training time (s) on an L40S GPU.}
        \label{tab:supp-ablation-per-scene}
        \end{table*}

\clearpage
\newlength{\eqfullcol}\setlength{\eqfullcol}{\dimexpr(\textwidth-4pt)/3\relax}
\newcommand{\eqfullimg}[1]{\raisebox{-0.5\height}{\includegraphics[width=\eqfullcol]{figures/equal_mem_renders/#1.jpg}}}
\newcommand{\eqfullrow}[7]{%
  \eqfullimg{#1_full_gt} & \eqfullimg{#1_full_fastergs} & \eqfullimg{#1_full_ours} \\
  PSNR / SSIM / LPIPS & $#2$\,dB / $#3$ / $#4$ & $#5$\,dB / $#6$ / $#7$ \\ \noalign{\vskip 3pt}}

\begin{figure*}[t]
\centering\scriptsize\setlength{\tabcolsep}{1pt}\renewcommand{\arraystretch}{1.0}%
\begin{tabular}{@{}ccc@{}}
  Ground truth & {[Adam + bwd.\ fusion]}, 10.0M & {[Ours + bwd.\ fusion]}, 25.3M \\
  \eqfullrow{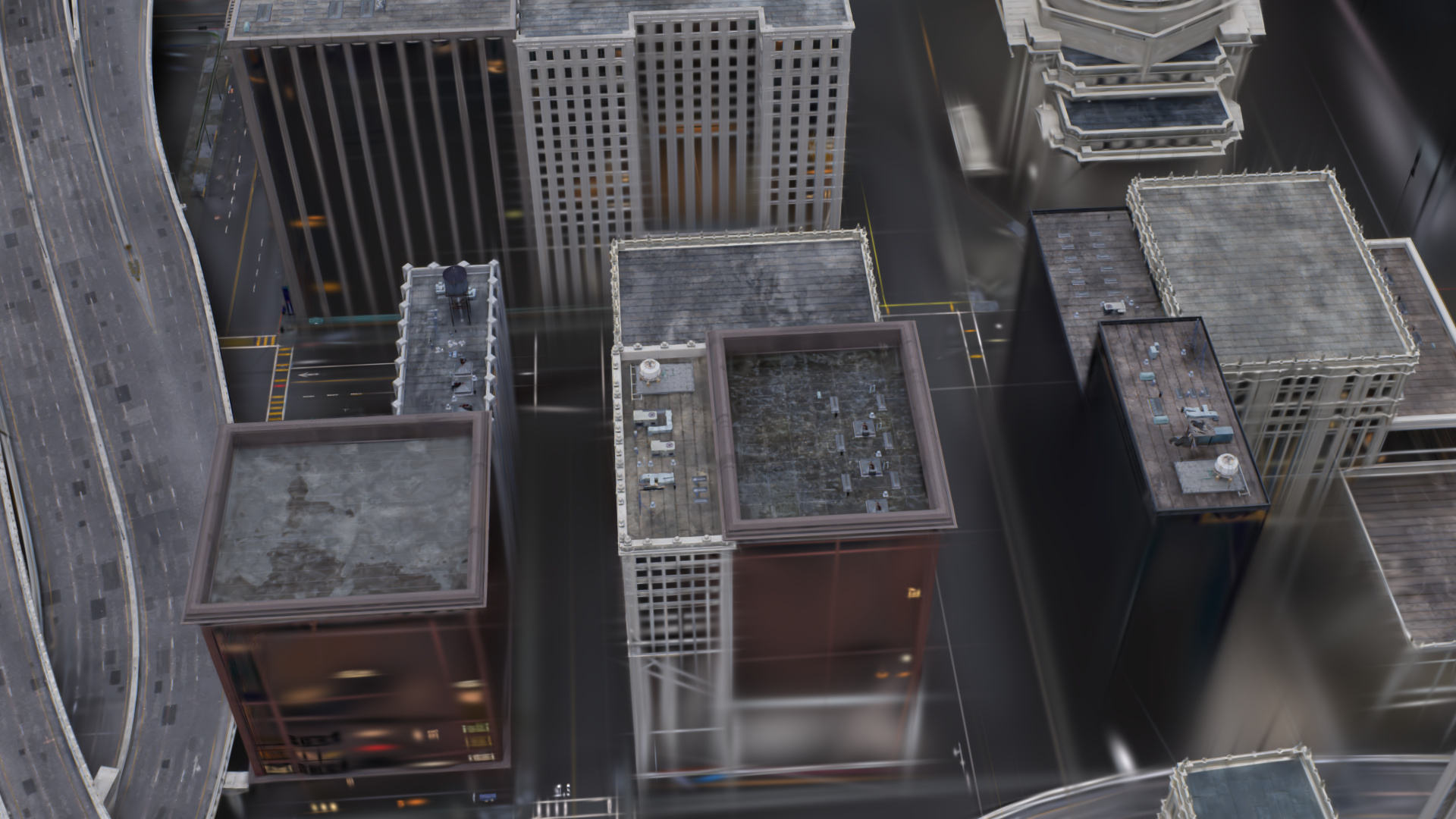}{25.06}{0.761}{0.415}{28.36}{0.845}{0.311}
  \eqfullrow{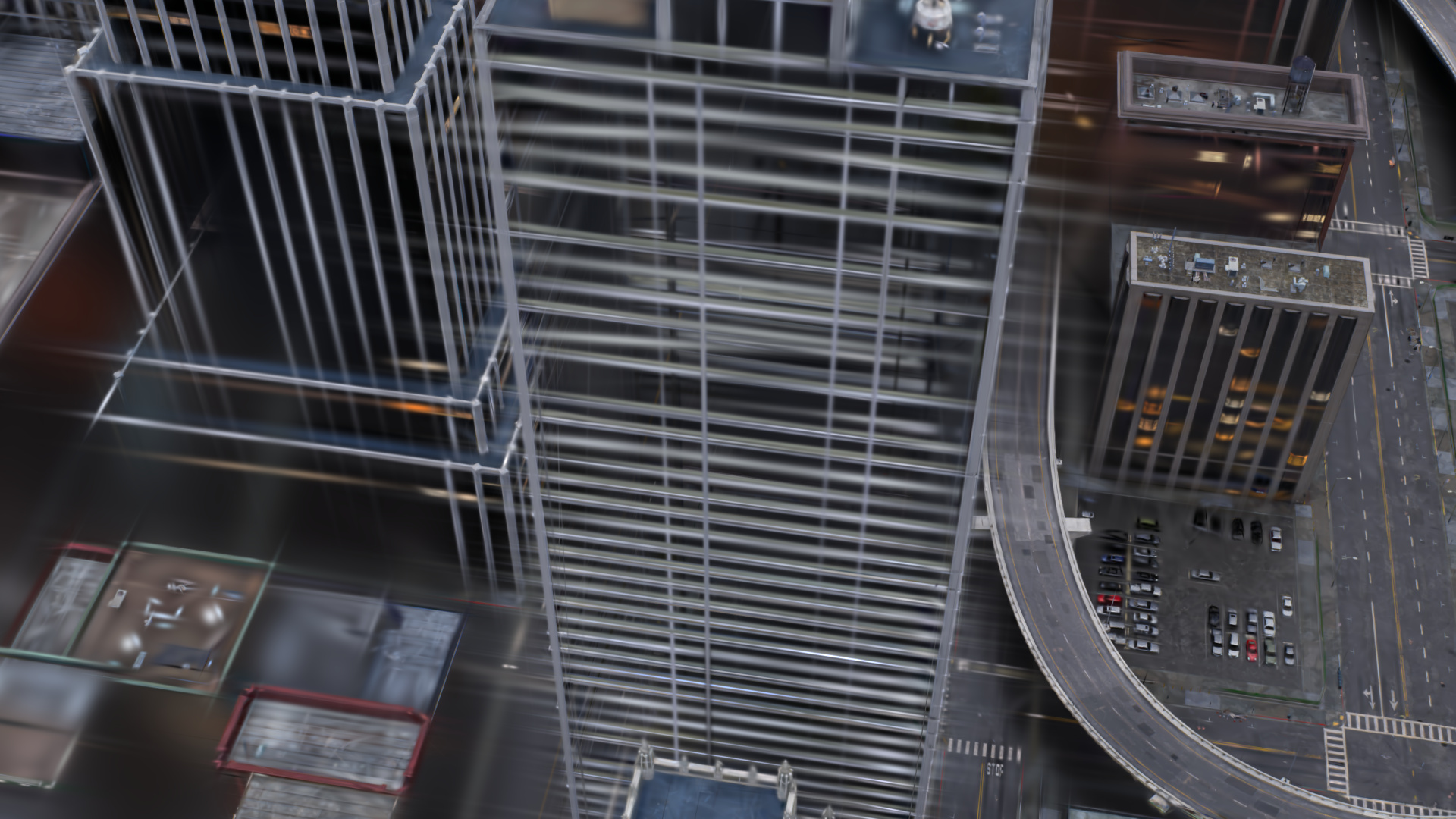}{20.25}{0.635}{0.458}{23.31}{0.739}{0.343}
  \eqfullrow{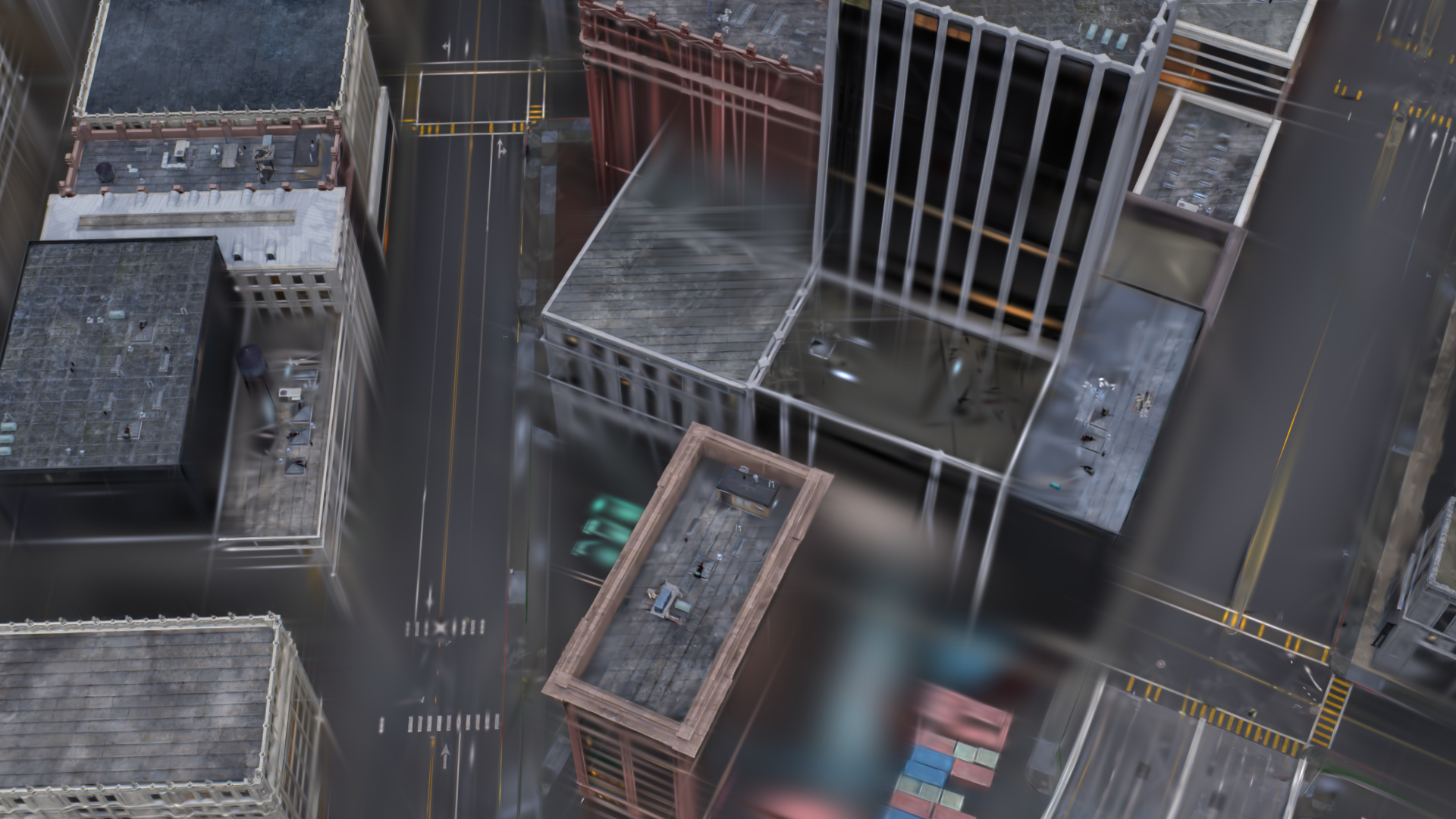}{24.10}{0.731}{0.457}{27.06}{0.813}{0.358}
  \eqfullrow{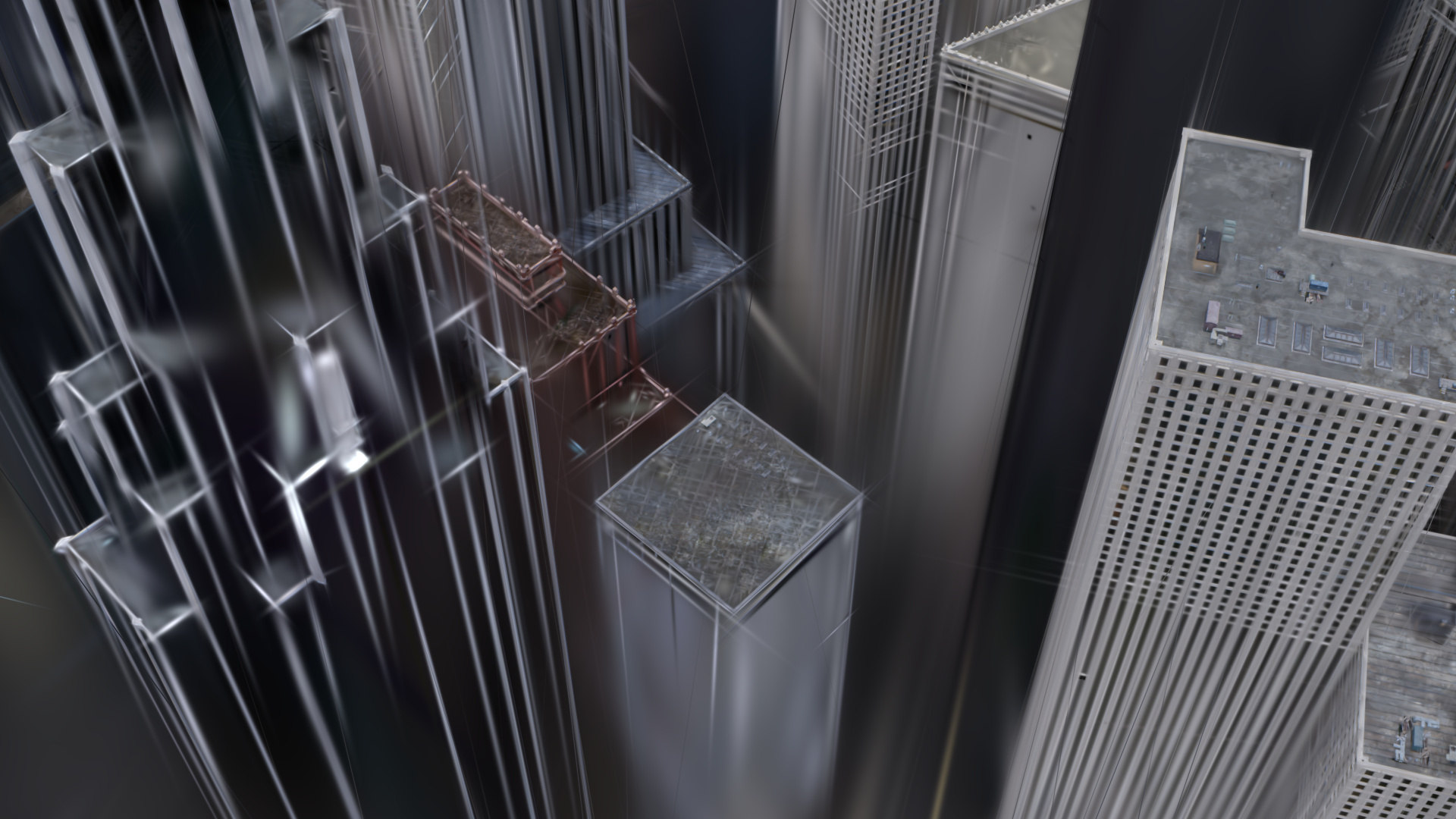}{18.49}{0.520}{0.537}{21.46}{0.666}{0.407}
  \eqfullrow{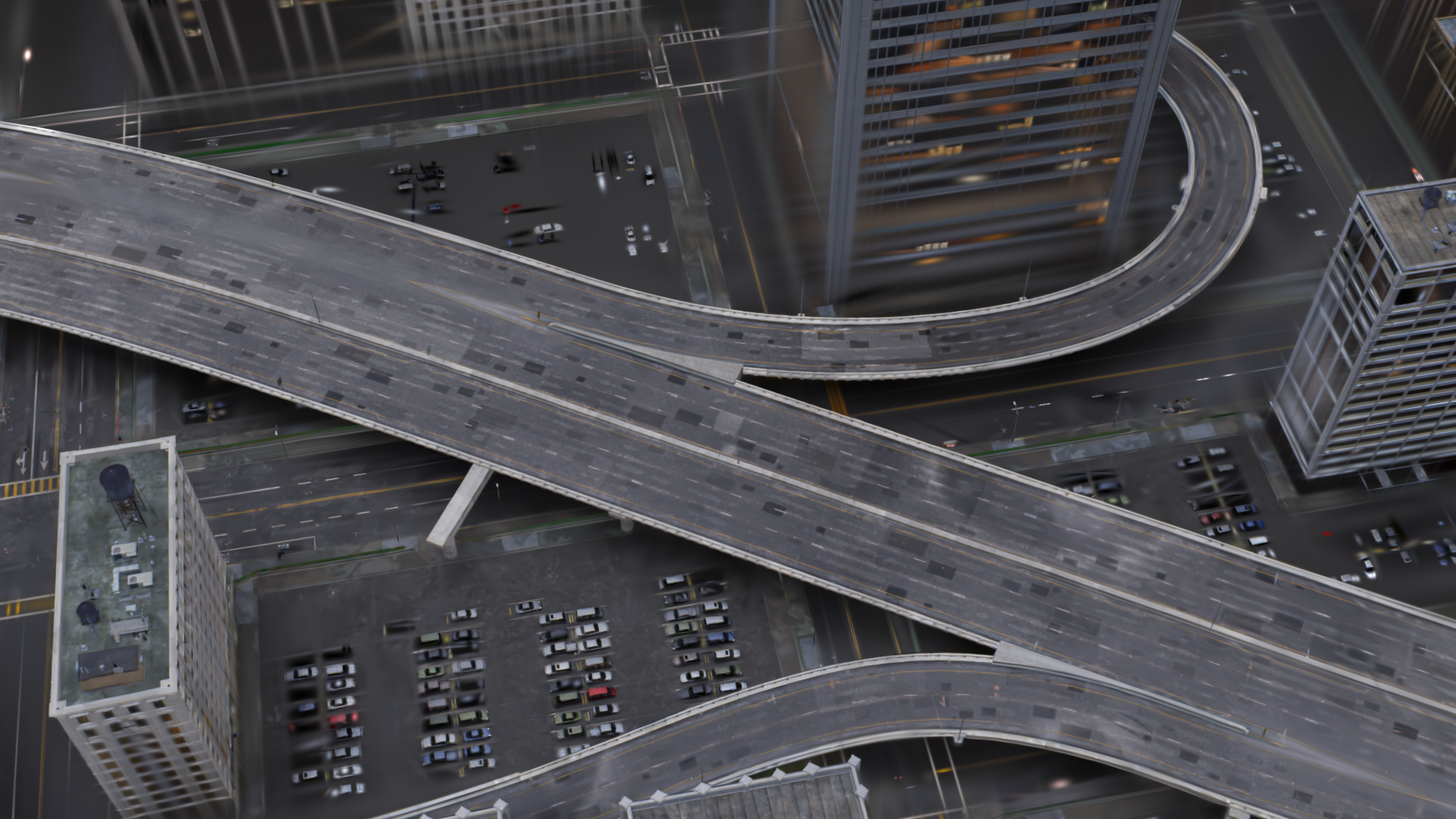}{26.05}{0.799}{0.364}{28.26}{0.874}{0.245}
\end{tabular}
\caption{Additional views of the experiment we show in the
teaser of the main paper. Since at equal training memory, we can fit $2.5
\times$ more Gaussians, we can recover a lot more details of the original scene
compared to Adam, whose lower primitive count cannot represent them. The PSNR, SSIM and LPIPS of each image are listed below it.}
\label{fig:equal_mem_renders}
\end{figure*}

\newlength{\rendercol}\setlength{\rendercol}{\dimexpr(\textwidth-22pt)/3\relax}
\newcommand{\renderimg}[1]{\raisebox{-0.5\height}{\includegraphics[width=\rendercol]{figures/renders/#1.jpg}}}
\newcommand{\renderrow}[9]{%
  \raisebox{-0.5\height}{\rotatebox{90}{\shortstack{#1\\\textit{#2}}}} &
  \renderimg{#2_#3_gt} & \renderimg{#2_#3_adam} & \renderimg{#2_#3_ours48full} \\
  & PSNR / SSIM / LPIPS & $#4$\,dB / $#5$ / $#6$ & $#7$\,dB / $#8$ / $#9$ \\ \noalign{\vskip 3pt}}

\begin{figure*}[p]
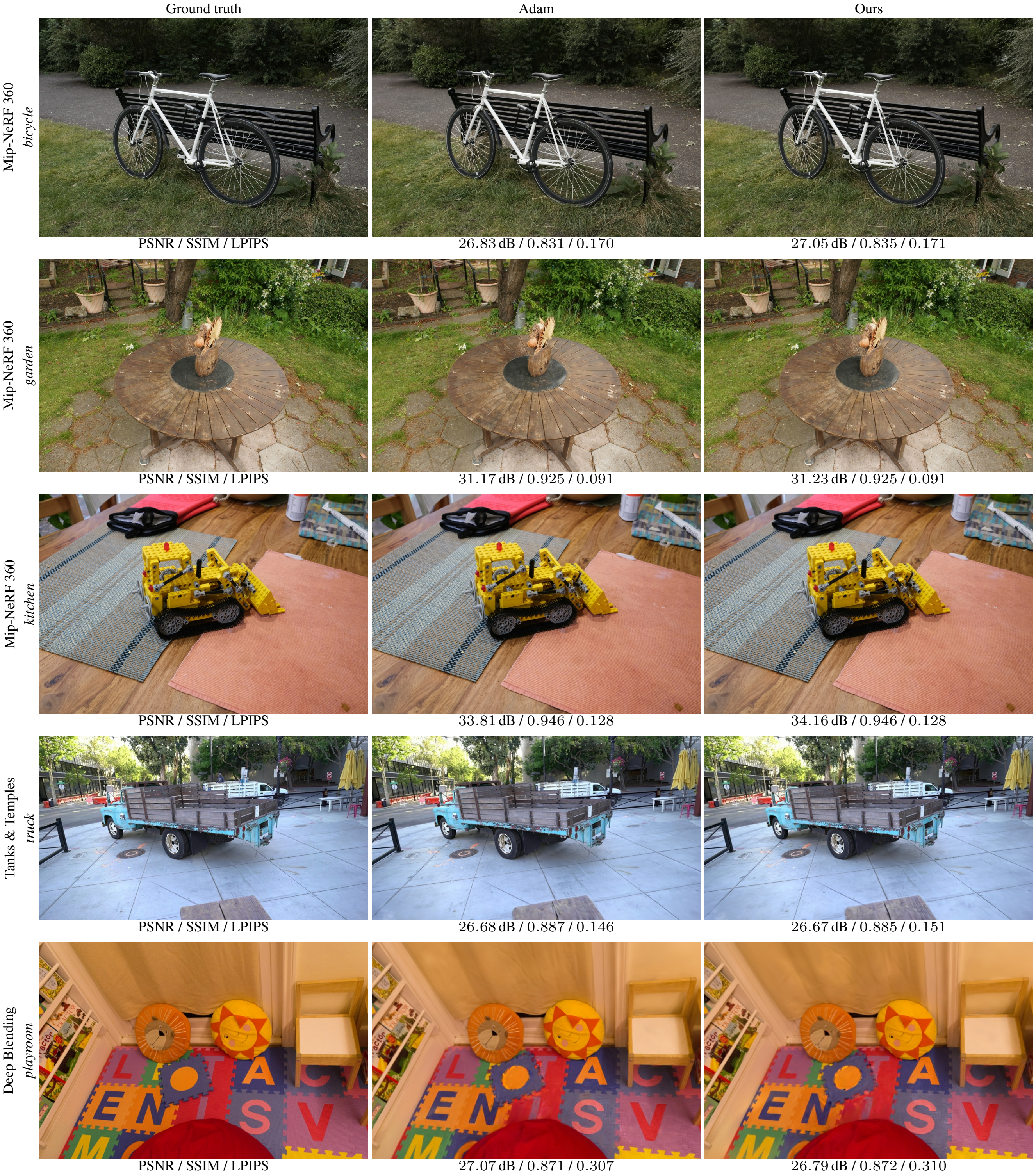

\centering\scriptsize\setlength{\tabcolsep}{1pt}\renewcommand{\arraystretch}{1.0}%
\begin{tabular}{@{}c*{3}{c}@{}}
  & Ground truth & Adam & Ours \\
  \renderrow{Mip-NeRF 360}{bicycle}{00012}{26.83}{0.831}{0.170}{27.05}{0.835}{0.171}
  \renderrow{Mip-NeRF 360}{garden}{00017}{31.17}{0.925}{0.091}{31.23}{0.925}{0.091}
  \renderrow{Mip-NeRF 360}{kitchen}{00017}{33.81}{0.946}{0.128}{34.16}{0.946}{0.128}
  \renderrow{Tanks \& Temples}{truck}{00016}{26.68}{0.887}{0.146}{26.67}{0.885}{0.151}
  \renderrow{Deep Blending}{playroom}{00014}{27.07}{0.871}{0.307}{26.79}{0.872}{0.310}
\end{tabular}
\caption{Test views rendered by the models trained with Adam and
Ours on scenes of Mip-NeRF 360~\cite{barron2022mipnerf360}, Tanks \&
Temples~\cite{Knapitsch2017} and Deep Blending~\cite{DeepBlending2018}. The PSNR, SSIM and LPIPS of each image are listed below it. Both optimizers use 3DGS-MCMC with the same primitive budget.}
\label{fig:renders}
\end{figure*}

\newlength{\clmcol}\setlength{\clmcol}{\dimexpr(\textwidth-14pt)/3\relax}
\newcommand{\clmimg}[1]{\raisebox{-0.5\height}{\includegraphics[width=\clmcol]{figures/clm_renders/#1.jpg}}}
\newcommand{\clmrow}[7]{%
  \raisebox{-0.5\height}{\rotatebox{90}{$#1$M}} &
  \clmimg{#1m_gt} & \clmimg{#1m_stock} & \clmimg{#1m_ours48} \\
  & PSNR / SSIM / LPIPS & $#2$\,dB / $#3$ / $#4$ & $#5$\,dB / $#6$ / $#7$ \\ \noalign{\vskip 3pt}}

\begin{figure*}[t]
\centering\scriptsize\setlength{\tabcolsep}{1pt}\renewcommand{\arraystretch}{1.0}%
\begin{tabular}{@{}c*{3}{c}@{}}
  & Ground truth & Sel.\ Adam & Ours \\
  \clmrow{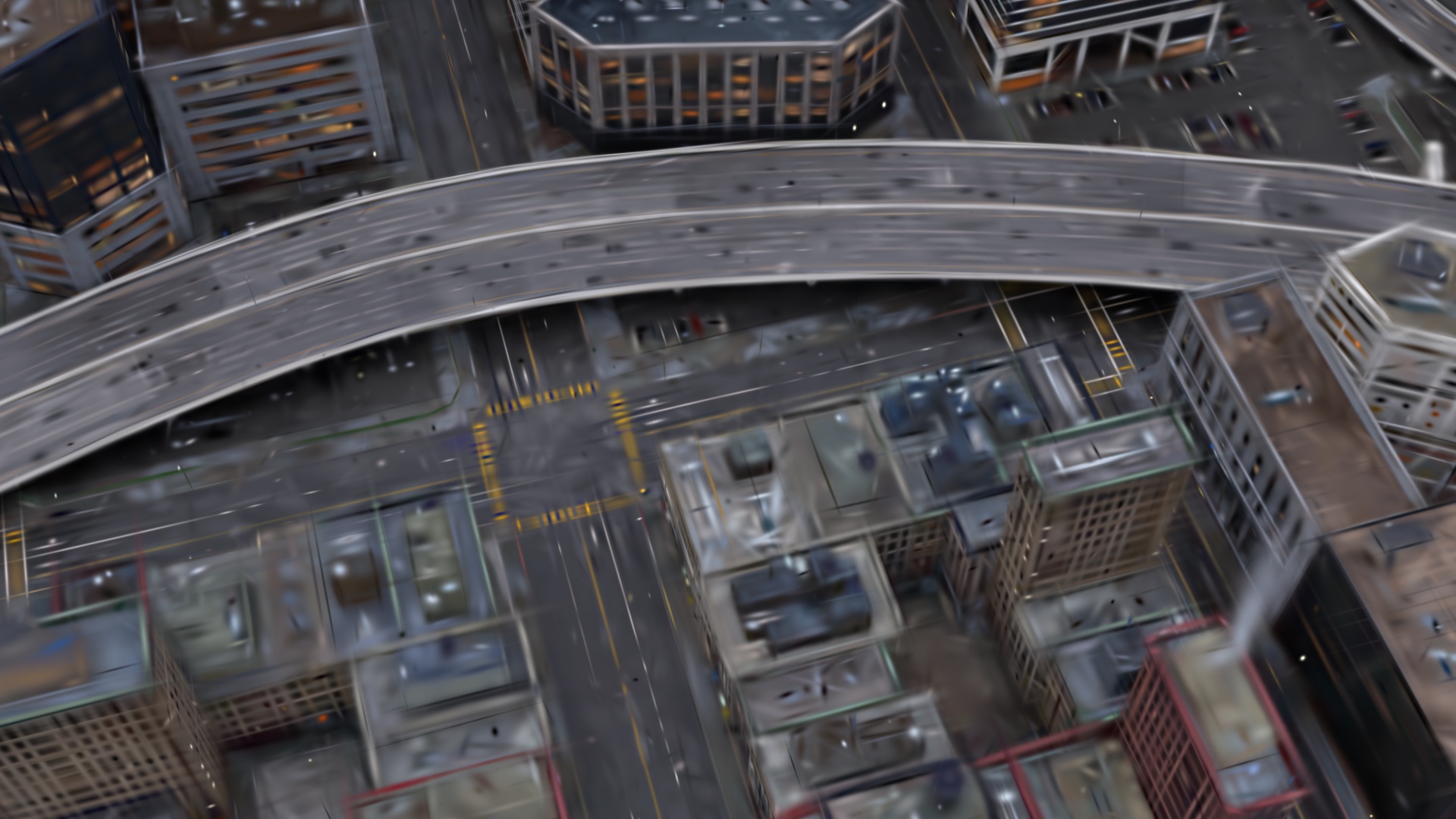}{24.85}{0.689}{0.502}{24.88}{0.688}{0.501}
  \clmrow{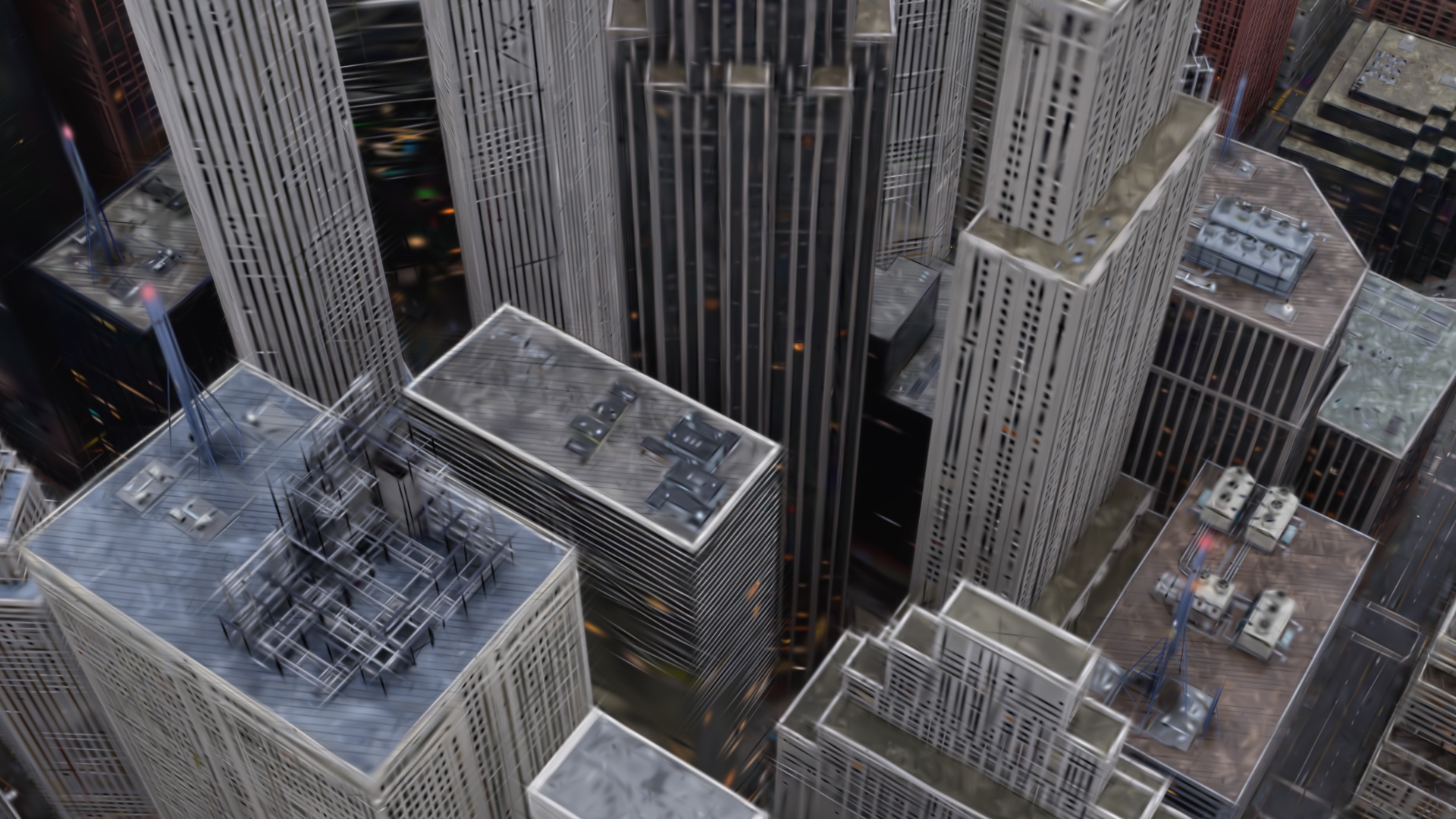}{22.90}{0.650}{0.435}{22.97}{0.650}{0.438}
  \clmrow{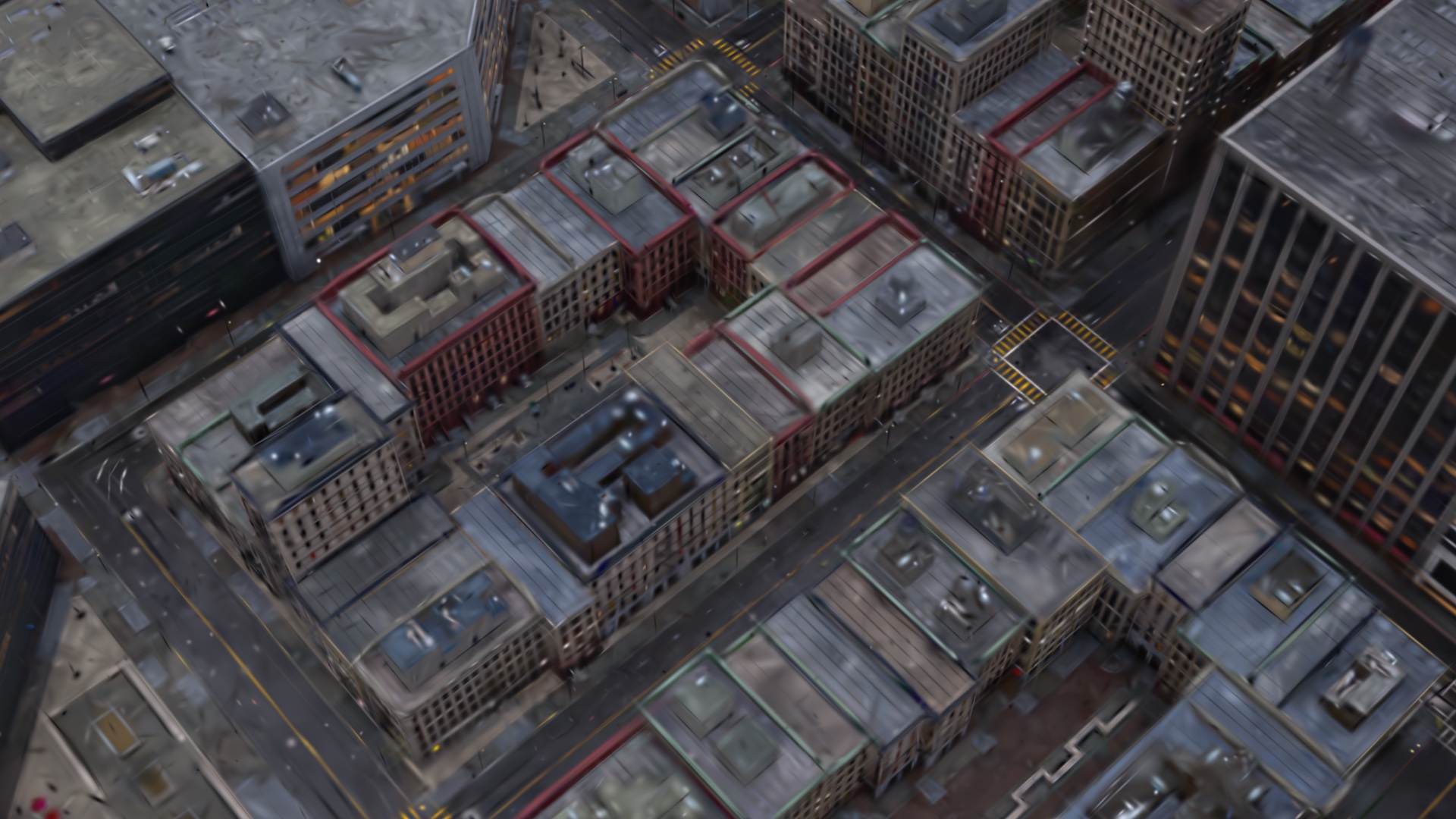}{24.79}{0.681}{0.465}{24.65}{0.676}{0.463}
\end{tabular}
\caption{Test views of MatrixCity (big city, aerial views)~\cite{li2023matrixcity}
rendered by CLM~\cite{zhao2025clm} trained with Selective Adam (Sel. Adam) and with our optimizer (Ours).
Each row shows a different view, rendered by the models trained with different primitive budgets. The PSNR, SSIM and LPIPS of each image are listed below it.}
\label{fig:clm_renders}
\end{figure*}

\end{document}